\documentclass[10pt,twocolumn,letterpaper]{article}

\usepackage[pagenumbers]{wacv}
\newcommand{\MAST}{\bf {MAST}}
\usepackage{xr-hyper}

\definecolor{wacvblue}{rgb}{0.21,0.49,0.74}

\usepackage{multirow}
\usepackage{subcaption}

\usepackage{arydshln}

\usepackage[
    pagebackref,
    breaklinks,
    colorlinks,
    allcolors=wacvblue
]{hyperref}

\title{MAST: Mask-Guided Attention Control for Training-Free \\
Regional-Multi Style Transfer}

\author{
Dongkyung Kang\textsuperscript{*},
Jaeyeon Hwang\textsuperscript{*},
Junseo Park\textsuperscript{*},
Minji Kang\textsuperscript{*},
Yeryeong Lee\textsuperscript{*},\\
Beomseok Ko,
Hanyoung Roh,
Jeongmin Shin,
Hyeryung Jang\textsuperscript{$\dagger$}\\
Department of Computer Science \& Artificial Intelligence, Dongguk University\\
{\small
\textsuperscript{*}Equal contribution.
\quad
\textsuperscript{$\dagger$}Corresponding author.
}
}

\begin{document}

\twocolumn[{%
\renewcommand\twocolumn[1][]{#1}%
\maketitle

\centering
\includegraphics[width=0.93\textwidth]{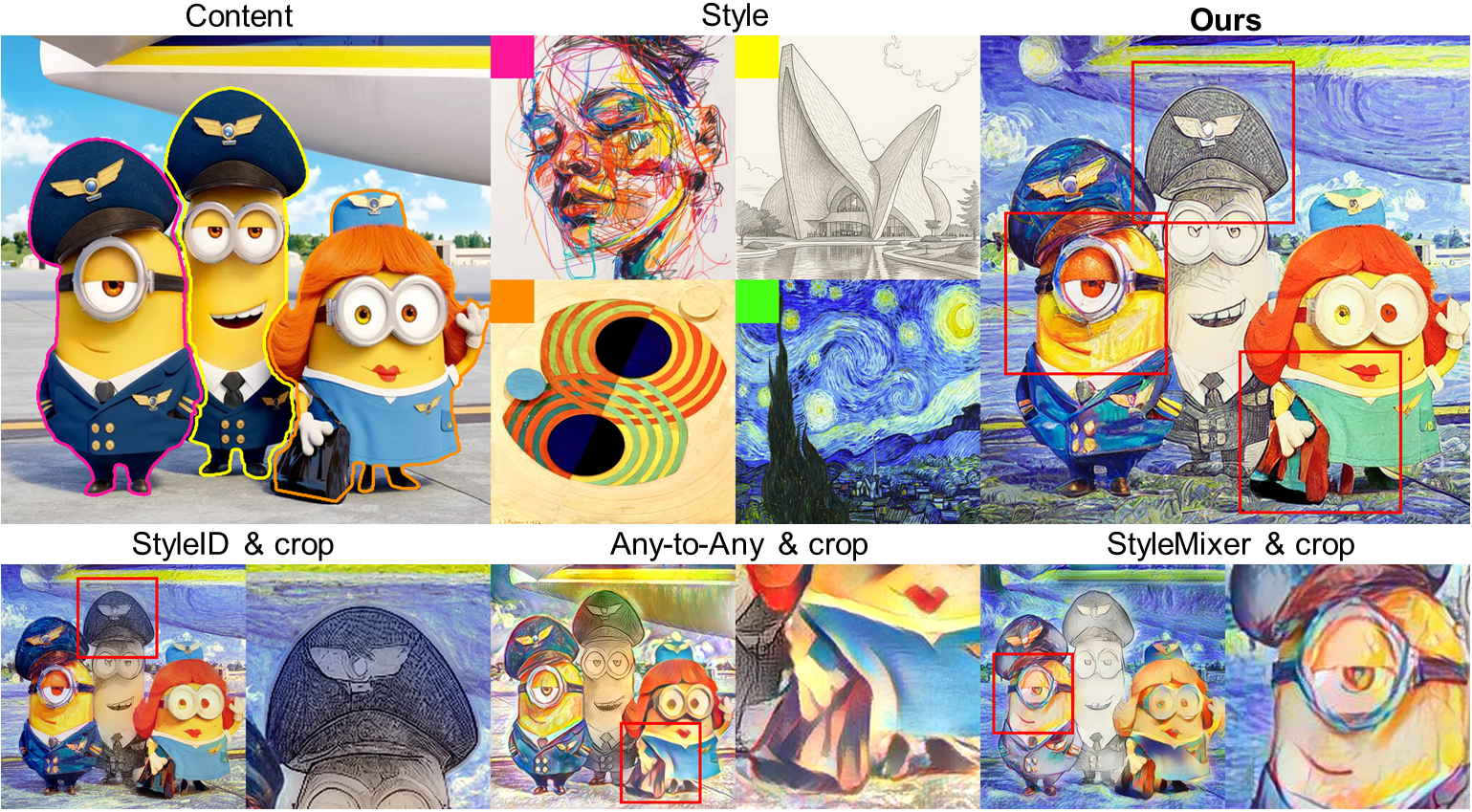}

\captionsetup{hypcap=false}
\captionof{figure}{
Regional-multi style transfer in the four-style setting.
{\MAST} (Row 1) preserves fine content details while successfully
transferring all four styles to their designated regions,
whereas representative baselines from
{\bf G-S} (StyleID~\cite{StyleID}),
{\bf R-S} (Any-to-Any~\cite{Any-to-any}), and
{\bf R-M} (StyleMixer~\cite{StyleMixer}) categories (Row 2)
exhibit detail degradation or unsuccessful transfer,
as highlighted by the crops.
Color-coded boxes on the style images and corresponding outlines
on the content image denote target spatial masks
(\textcolor{green}{green}: background).
}

\label{fig:teaser}

\vspace{0.5em}
}]

% =========================
% Main Paper
% =========================

\begin{abstract}
Style transfer applies the appearance of a reference image to a content image while preserving its spatial structure. 
Recent diffusion-based methods achieve strong stylization but typically assume a single global style. 
We instead consider regional-multi style transfer, which assigns multiple references to user-specified regions of a content image. 
Extending them to this setting reveals two coupled shared-attention issues: ambiguous mass allocation among content and style partitions and 
degraded selectivity as more styles are jointly normalized, while style aggregation at the attention output further suppresses fine details.

We propose {\MAST} (Mask-Guided Attention Control for Training-Free Regional-Multi Style Transfer), a unified attention-control framework for frozen diffusion models. Logit-level Attention Mass Allocation enforces mask-derived partition masses, Sharpness-aware Temperature Scaling adaptively restores selectivity, and Discrepancy-aware Detail Injection recovers high-frequency content. 
{\MAST} jointly processes all style--mask pairs in a single denoising pass without training, optimization, or post-hoc composition. 
Across two to five styles, {\MAST} achieves the best average ArtFID, FID, and R-FID among all baselines, demonstrating regional style fidelity, content preservation, and scalability.
\vspace{-0.5em}

\end{abstract}

\section{Introduction} 
\label{sec:intro}

Image style transfer aims to stylize a content image according to the appearance of a reference image while preserving its semantic structure and spatial layout~\cite{StyCNN,Adain,universal,learned,StyTr2}. Recent diffusion-based approaches have substantially advanced arbitrary style transfer by leveraging the internal representations of pretrained diffusion models~\cite{inST,zstar, StyleID,DiffStyle,diffusest,stylessp}.
However, practical editing often requires region-wise control, where different reference styles are assigned to distinct objects or background regions. We refer to this setting as \emph{regional-multi style transfer}.

We characterize style transfer along two axes: the number of style references (single vs. multiple)~\cite{wang2024multisource, ruta2025leveraging} and the spatial extent of stylization (global vs. regional), yielding four settings: global-single (\textbf{G-S}), global-multi (\textbf{G-M}), regional-single (\textbf{R-S}), and regional-multi (\textbf{R-M}). 
We focus on regional-multi style transfer, 
where each style reference is 
paired with a spatial mask. Unlike prior methods based on automatic semantic 
matching~\cite{StyleMixer,StyleGallery} or post-hoc spatial 
composition~\cite{ye2023multistyle}, we realize explicit style--region 
correspondence through training-free diffusion attention control. 
The goal is to faithfully reproduce each assigned style in its designated region while preserving content structure and maintaining visually coherent boundaries.

Prior regional-multi methods coordinate styles outside shared attention
normalization through semantic matching or spatial composition, controlling which style applies where but not how content and style partitions compete once jointly normalized.
Viewing this shared-attention perspective reveals two coupled limitations. 
First, {\em partition-allocation ambiguity}: spatial masks specify where each style should be applied but do not determine how attention is distributed among the content and style partitions, causing cross-region interference or weak regional stylization (\cref{fig:analysis_lama}). 
Second, {\em attention-sharpness degradation}: integrating multiple style sources can flatten the joint attention distribution, 
weakening distinctive style retrieval and
diluting textures (\cref{fig:analysis_sts}). 
Although reduced Softmax selectivity has been studied generally~\cite{velivckovic2024softmax}, its interaction with regional allocation in multi-style diffusion remains underexplored, making joint control of allocation and sharpness essential.

Beyond these attention-control challenges, content details can be degraded at
the attention output stage. Although standard query blending~\cite{StyleID}
stabilizes spatial structure, subsequent style-feature aggregation can
suppress content-specific boundaries and fine structures
~\cite{StyleFM,diffusest,res-based2}. This motivates a complementary
output-stage mechanism for detail recovery.

Building on this analysis, we propose {\bf {\MAST}} ({\bf M}ask-Guided {\bf A}ttention Control for Training-Free Regional-Multi {\bf S}tyle {\bf T}ransfer), a training-free framework comprising two core attention-control modules and one complementary detail-preservation module. 
{\MAST} jointly integrates multiple style-mask pairs within a single denoising process of a frozen diffusion model, without additional training or per-sample optimization. 
The key insight is that both limitations admit a pre-softmax solution: mass allocation via closed-form logit biases, and selectivity restoration via adaptive temperature scaling anchored to content self-attention.

At the mass allocation stage, Logit-level Attention Mass Allocation (LAMA) enforces mask-derived query-wise partition masses through closed-form logit biases, suppressing unrelated partitions. 
At the softmax normalization stage, Sharpness-aware Temperature Scaling (STS)
restores selectivity by adapting the temperature to the sharpness gap from content self-attention across heads, layers, and timesteps.
At the attention output stage, Discrepancy-aware Detail Injection (DDI) restores high-frequency content based on the feature discrepancy between
stylization and content.

Averaged across the two to five styles, {\MAST} achieves the best ArtFID ($14.928$, $7.1\%$ lower than the next-best baseline), FID ($8.798$), and R-FID ($16.290$) among all baselines.  
Fig.~\ref{fig:exp_base} further 
shows that {\MAST} preserves
fine details, coherent boundaries, and regional style fidelity where baselines from all three categories (\textbf{G-S}, \textbf{R-S}, \textbf{R-M}) exhibit boundary
artifacts, style bleeding, or diluted textures.

Our main contributions are summarized as follows:
\begin{itemize}
\item We introduce a training-free framework for regional-multi style transfer with user-specified style--region correspondence, identifying two coupled shared-attention challenges: partition-allocation ambiguity and attention-sharpness degradation.

\item We propose Logit-level Attention Mass Allocation (LAMA), operating at the mass allocation stage, 
which enforces mask-derived target masses at each query through closed-form logit biases.

\item We propose Sharpness-aware Temperature Scaling (STS), operating at the softmax normalization stage,  
which adaptively
restores selectivity using the sharpness gap from
content self-attention.

\item We propose Discrepancy-aware Detail Injection (DDI), operating at the attention output stage, which restores high-frequency information through feature-level discrepancy-weighted injection.

\end{itemize}

\section{Related work}

\noindent \textbf{Global style transfer.} 
Global artistic style transfer applies a reference style throughout the
content image without spatial partitioning. 
Early CNN-based methods matched perceptual features and global style statistics~\cite{StyCNN, universal, learned}, with AdaIN~\cite{Adain} enabling arbitrary style transfer via channel-wise statistics alignment; Transformers later improved
long-range content--style interactions~\cite{StyTr2,FineGrainedST}. 
Diffusion-based approaches advanced through style-specific adaptation~\cite{inST, Stylediffusion, Consislora} or training-free manipulation of internal representations~\cite{DiffStyle, TFLDST, AttenST, diffusest, stylessp}. 
Notably, StyleID~\cite{StyleID} substitutes self-attention keys/values while preserving content via query preservation, and $Z^*$~\cite{zstar} controls style injection through attention reweighting. 
Recent works exploit frequency-aware injection~\cite{StyleFM, res-based2} and diffusion-feature semantic correspondence~\cite{CoCoDiff}. These {\em global-single} methods apply one style throughout the image, whereas {\em global-multi} extensions~\cite{yu2025multisource, ruta2025leveraging, wang2024multisource, SCAdapter} fuse multiple style references into a global style representation without
explicit correspondence to distinct spatial regions.

% \noindent \textbf{Regional-single style transfer.} 
\noindent \textbf{Regional style transfer.} 
Regional style transfer localizes stylization through masks, semantic regions, or user interaction.
%, enabling spatial control that global methods do not provide. 
{\em Regional-single} methods enable pixel-level control via test-time training~\cite{kim2024controllable}, transfer a selected style region to a designated content region (Any-to-Any~\cite{Any-to-any}), improve boundary handling through internal feature blending (BPC~\cite{BPC}), or supervise style-token attention with
object masks during training~\cite{chen2026regionroute}. 
These methods primarily localize one style at a time and do not address
simultaneous interactions among multiple style representations.

% \noindent \textbf{Regional-multi style transfer.} 
{\em Regional-multi} methods assign multiple styles to different regions via automatic semantic matching (StyleMixer~\cite{StyleMixer}, StyleGallery~\cite{StyleGallery}), image-space blending of independently stylized instances~\cite{ye2023multistyle}, or per-style customized models~\cite{li2026efficient}. 
These external mechanisms do not explicitly regulate competition among content and style
partitions within shared attention. 
{\MAST} instead processes all style/content representations jointly within shared attention, directly regulating partition allocation and sharpness under user-specified explicit correspondence, without model specialization or training.

\section{Method}

\begin{figure}[t!]
    \centering

    \begin{subfigure}[t]{\linewidth}
        \centering
        \includegraphics[width=\linewidth]{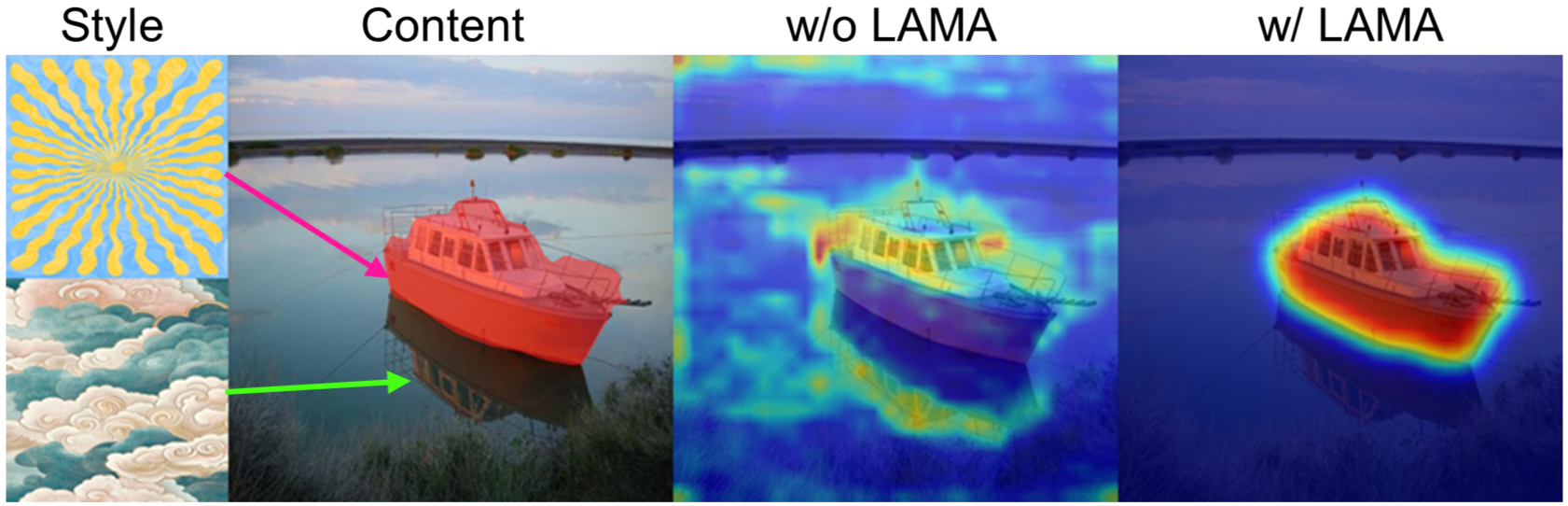}
        \caption{
        Partition-allocation ambiguity and LAMA
        }
        \label{fig:analysis_lama}
    \end{subfigure}

    \vspace{0.1cm}

    \begin{subfigure}[t]{\linewidth}
        \centering
        \includegraphics[width=\linewidth]{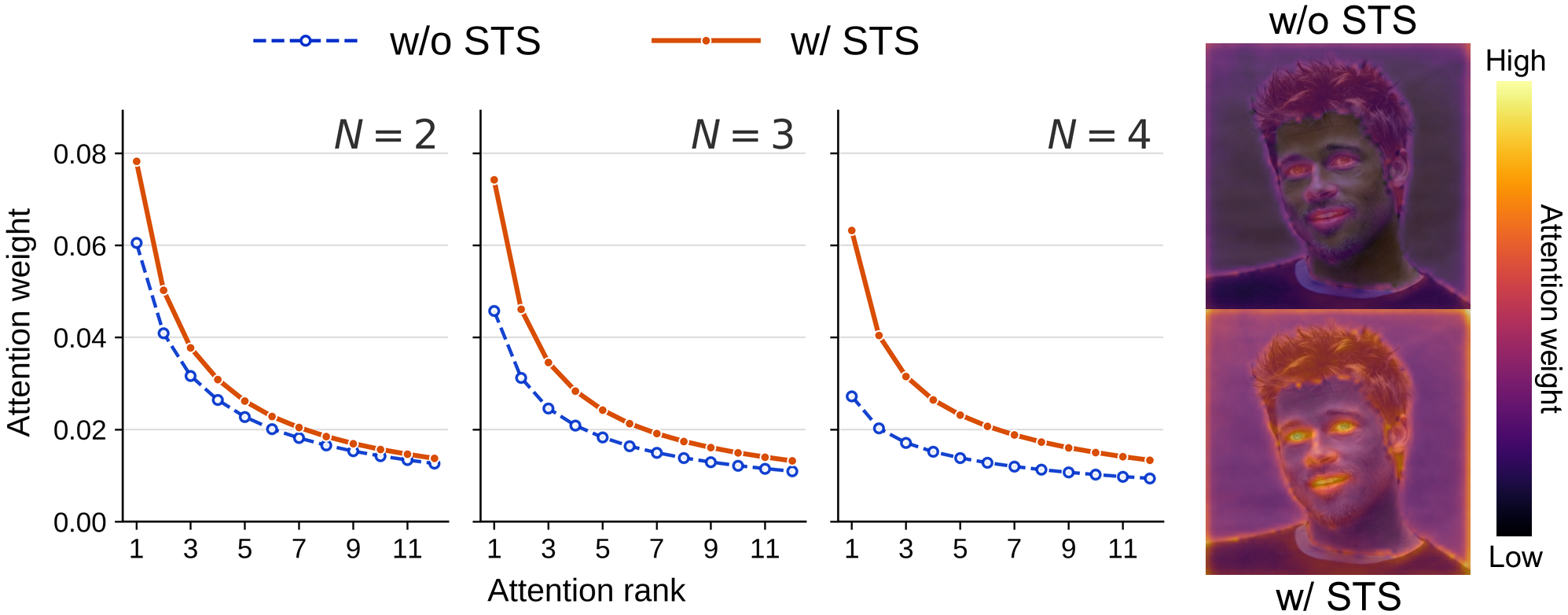}
        \caption{
         Attention-sharpness degradation and STS
        }
        \label{fig:analysis_sts}
    \end{subfigure}

    \caption{
    The two shared-attention failure modes and their corrections. 
    (A) Under shared attention, the attention mass assigned to the upper style reference spreads beyond the target region; LAMA focuses it on the mask. (B) Rank-wise attention becomes flatter as the number of style references $N$ increases, whereas STS preserves selectivity (left); the corresponding sharpness maps (right) further illustrate this improvement.
    }
    \label{fig:analysis}
    \vspace{-0.5cm}
\end{figure}

\begin{figure*}[t!]
    \centering
    \includegraphics[width=\textwidth]{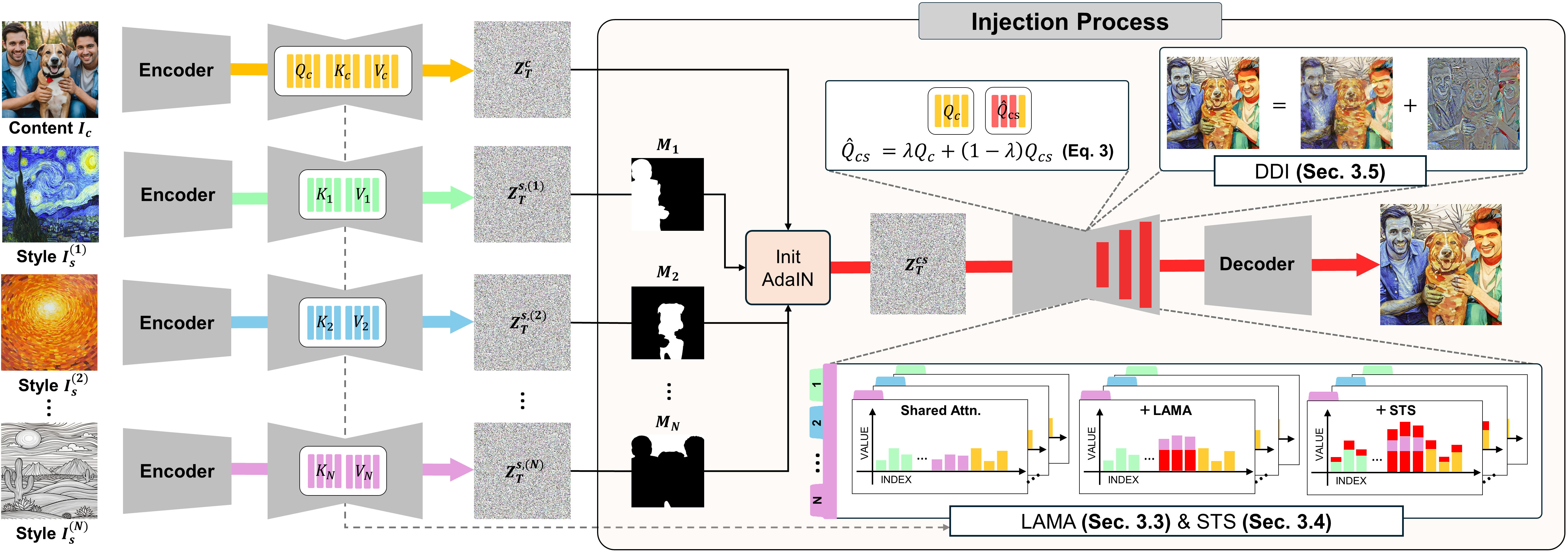}
    \caption{
    Overview of the {\MAST} framework. 
    (Left) The stylization pipeline employs DDIM inversion and AdaIN-based initialization, with style injected through attention modulation during denoising. 
    (Right) The attention control framework comprises two core attention-control modules - LAMA (mass allocation) and STS (attention sharpening) - and one complementary detail-preservation module (DDI). The bar charts illustrate attention allocation progressively reshaped by LAMA and sharpened by STS. 
    }
  \label{fig:overview}
  \vspace{-0.2cm}
\end{figure*}

\subsection{Analysis of regional-multi style transfer}
\label{sub:analysis}

Training-free diffusion style transfer injects reference appearance through
self-attention manipulation. StyleID~\cite{StyleID} substitutes content
keys and values with style features, whereas $Z^*$~\cite{zstar} jointly
normalizes style cross-attention and content self-attention. We adopt the shared-normalization structure of $Z^*$ and extend it
to multiple style partitions while retaining the content partition.

Let $Q_{cs}$ and $Q_c$ denote the content-stylized and content queries, and let $(K_c, V_c)$ and $(K_i, V_i)$ denote the content and $i$-th style key-value pairs. 
The style and content logits are $L_i = Q_{cs} K_i^\top / \sqrt{d}$ and
$L_c = Q_c K_c^\top / \sqrt{d}$, where $d$ is the head dimension. 
Extending to multiple styles, all partitions are concatenated and normalized jointly:
\begin{align}
A_\text{concat}
=
\operatorname{Softmax}
\left(
[L_1,\ldots,L_N,L_c]
\right).
\label{eq:concat-att}
\end{align} 
For clarity, \cref{eq:concat-att} uses $Q_{cs}$; the actual framework replaces it with the content-anchored query introduced in \cref{sec:overview}. This shared formulation exposes two coupled attention-control problems.

\noindent {\bf Partition-allocation ambiguity.} 
The mask $M_i$ specifies the spatial support of style $i$, but does not
appear in \cref{eq:concat-att}, so partition masses depend only on raw logits.
Consequently, a style can receive excessive mass outside its designated
region or insufficient mass inside it. As shown in \cref{fig:analysis_lama}, the attention mass assigned to the upper style
reference
spreads beyond its assigned object region without allocation control, whereas LAMA aligns it with
the target mask. Regional control therefore requires translating spatial
correspondence into explicit query-wise constraints on partition mass.

\noindent {\bf Attention-sharpness degradation.}
Adding style references increases the number of competing keys, reducing
selectivity when logit separation does not increase accordingly; weak
cross-image query--key correlations can further amplify this effect. As shown in \cref{fig:analysis_sts}, the rank-wise attention weights
become progressively flatter as more styles are added, whereas STS maintains
a more selective distribution with stronger emphasis on highly ranked keys.
Since the sharpness loss varies across heads, layers,
timesteps, and inputs, its correction must be adaptive.

These observations motivate two coupled controls: LAMA regulates mask-guided
partition allocation, while STS adaptively restores attention selectivity.
DDI complements them by recovering high-frequency content details suppressed
during output-stage feature aggregation.

\subsection{Overall framework}
\label{sec:overview}

Given a content image $I_c$ and $N$ style--mask pairs $\{(I_s^{(i)}, M_i)\}_{i=1}^N$, {\MAST} jointly injects all reference
styles within a single denoising process, where each $M_i$ specifies the target
region of style $i$. 
As illustrated in \cref{fig:overview}, DDIM inversion~\cite{ddim}
produces $z_T^c$ from $I_c$ and $z_T^{s,(i)}$ from each style image. 
Defining the unassigned content region as
$M_c = 1-\sum_{i=1}^{N}M_i$, we initialize the stylization latent using 
region-wise AdaIN~\cite{Adain}:
\begin{align} \label{eq:regional_adain}
    z_T^{cs}
    =
    M_c \odot z_T^c
    +
    \sum_{i=1}^{N}
    M_i \odot
    \operatorname{AdaIN}(z_T^c, z_T^{s,(i)}).
\end{align}
This initialization provides region-specific coarse appearance, while style features are subsequently injected through attention.

Following StyleID~\cite{StyleID}, we anchor the content-stylized  query to the content query:
\begin{align} \label{eq:query_blending}
    \widehat{Q}_{cs}
    =
    \lambda Q_c + (1-\lambda)Q_{cs},
\end{align}
where $\lambda\in[0,1]$ controls content anchoring. Accordingly, the
style logits $L_i$ in \cref{sub:analysis} use
$\widehat{Q}_{cs}$ in place of $Q_{cs}$, while the content self-attention
logits $L_c$ remain unchanged. 
This content-anchored query preserves the layout established by the AdaIN initialization while enabling region-specific style modulation through LAMA and STS. 

The subsequent attention-control and output-stage operations follow
\cref{fig:overview}: LAMA, STS, and DDI are applied sequentially at the mass allocation, softmax normalization, and attention output stages, as
detailed in \cref{sec:lama}--\cref{sec:ddi}. 
The entire framework
operates on a frozen diffusion model without fine-tuning or per-instance
optimization.

\subsection{Logit-level attention mass allocation}
\label{sec:lama}

As illustrated in \cref{fig:overview}, LAMA resolves
partition-allocation ambiguity at the mass allocation stage by enforcing
mask-derived attention masses at each spatial query before STS sharpening.
For a single attention head, $L_i$ and $L_c$ contain all spatial queries.
For one query, we denote the corresponding logit vectors by $\ell_i$ and
$\ell_c$ and omit the query index; $M_i$ denotes its mask value.
Given a maximum style-mass budget $\pi^\star$, the target masses are
\begin{align}
\pi_i^\star = \pi^\star M_i,
\qquad
\pi_c^\star = 1 - \sum_{k=1}^{N} \pi_k^\star.
\label{eq:lama_target}
\end{align}
Thus, a location fully assigned to style $i$ allocates mass $\pi^\star$
to that style, whereas an unassigned location retains the full content
mass. Continuous mask values interpolate between these cases.

Let $Z_i=\sum_j \exp(\ell_{i,j})$ and
$Z_c=\sum_j\exp(\ell_{c,j})$ denote the partition functions of
style $i$ and content, respectively, where $j$ indexes the keys.
Before LAMA, their style-to-content mass ratio is $Z_i/Z_c$. LAMA
corrects this ratio toward the mask-derived target
$\pi_i^\star/\pi_c^\star$ by adding a scalar bias $b_i$ to all key
logits in style partition $i$: the partition is increased when its
current ratio is below the target and decreased when it exceeds the
target. Matching the biased ratio $\exp(b_i)Z_i/Z_c$ to the target gives
\begin{align}
b_i
=
% \log\left(\frac{\pi_i^\star}{\pi_c^\star}\right)
\log\left( {\pi_i^\star} \big/{\pi_c^\star} \right)
+
\log Z_c
-
\log Z_i.
\label{eq:lama_bias}
\end{align}
The resulting joint logit vector is
\begin{align}
\ell_{\mathrm{LAMA}}
=
\left[
\ell_1+b_1,\ldots,\ell_N+b_N,\ell_c
\right].
\label{eq:lama_logits}
\end{align}
The content partition remains unbiased as the reference. Since $b_i$ is
shared across all keys in style partition $i$, it changes partition-level
competition while preserving intra-partition relative differences.
Further derivations and the selection of $\pi^\star$ are provided in
Appendix \cref{app:lama}.

\subsection{Sharpness-aware temperature scaling}
\label{sec:sts}

At the softmax normalization stage, STS restores joint-attention selectivity
by adapting a sharpening factor to the sharpness gap from content
self-attention. As discussed in \cref{sub:analysis}, concatenating more style partitions can flatten the distribution, making a fixed factor used in prior
work~\cite{StyleID} too weak under severe flattening or too strong for already
selective attention. As illustrated in \cref{fig:overview}, STS
sharpens the LAMA-adjusted distribution by concentrating attention on
dominant keys. It uses $\tau$ as a multiplicative inverse-temperature factor,
estimated head-wise at each controlled layer and timestep.

Stacking $\ell_{\mathrm{LAMA}}$ over the $N_q$ spatial queries yields the head-wise logit matrix $L_{\mathrm{LAMA}}$. For any such matrix $L$, the log peak probability of its $q$-th query row is
\begin{equation}
\log p_{\max}(L_{q,:})
=
\max_j L_{qj}
-
\log\sum_j\exp(L_{qj}).
\label{eq:log_pmax}
\end{equation}
We use $\log p_{\max}(L)$ to denote the mean of these scores over the $N_q$ query rows. 
Since each $p_{\max}(L_{q,:})\in(0,1]$, this average approaches zero for sharper attention and becomes more negative as attention flattens. 
This score is computed directly from the
logits without materializing an additional Softmax map.

With content self-attention as a reference independent of the number of style partitions, the sharpness gap is
\begin{equation}
\Delta
=
\log p_{\max}(L_c)
-
\log p_{\max}(L_{\mathrm{LAMA}}).
\label{eq:sharpness_gap}
\end{equation}
A positive $\Delta$ indicates a joint-attention sharpness deficit. The quadratic mapping $\tau=f(\Delta)$ produces a scalar factor applied to the LAMA-adjusted logits of the current head:
\begin{equation}
A_{\mathrm{STS}}
=
\operatorname{Softmax}
\left(
\tau L_{\mathrm{LAMA}}
\right).
\label{eq:sts_attention}
\end{equation}
Further
analysis of the $\Delta$-to-$\tau$ mapping and the distributional effects of
STS is provided in Appendix \cref{appendix:tau}--\cref{appendix:sts_interaction}.

\subsection{Discrepancy-aware detail injection}
\label{sec:ddi}

Even after LAMA and STS regulate attention allocation and sharpness, feature aggregation can suppress content-specific high-frequency details such as boundaries and fine structures. Content-anchored query blending~\cite{StyleID} stabilizes spatial layout but does not explicitly recover them. DDI therefore injects high-frequency content features according to the discrepancy between stylization and content residual updates. As illustrated in \cref{fig:overview}, this injection enhances fine structures suppressed during stylization.

Let $F_x$ denote the skip feature of the stylization path, $\Delta F_x$ its residual-branch update (the output added to $F_x$ within the residual block), and $F_c$ the residual-branch output of the corresponding content-path ResBlock. In our implementation, these features are obtained from the last ResBlock of each controlled decoder output block. Following prior residual-based approaches~\cite{diffusest,res-based2}, we extract the high-frequency component of $F_c$ using a Gaussian high-pass operator $\mathcal{H}_r$:
\begin{equation}
F_c^{\mathrm{high}} = \mathcal{H}_r(F_c).
\label{eq:high_frequency}
\end{equation}
The operator $\mathcal{H}_r$ applies a Gaussian high-pass mask via 2D Fourier transforms, and $r$ controls its normalized frequency scale. DDI determines the injection strength from the cosine discrepancy between $\Delta F_x$ and $F_c$:
\begin{equation}
\begin{aligned}
\omega &= 1-\cos(\Delta F_x,F_c),\\
\widehat{F}_x &= F_x+\Delta F_x+\omega F_c^{\mathrm{high}}.
\end{aligned}
\label{eq:ddi}
\end{equation}

In \cref{eq:ddi}, cosine similarity is computed jointly over the channel and spatial dimensions, yielding a scalar discrepancy weight $\omega \in [0,2]$ per sample at each DDI application. 
Larger discrepancy strengthens high-frequency compensation, preserving fine structures while limiting interference with transferred styles. Filter details are provided in Appendix \cref{appendix:high}.

\begin{table*}[t]
\centering

\scriptsize
\setlength{\tabcolsep}{2.5pt}
\setlength{\arrayrulewidth}{0.4pt}
\setlength{\dashlinedash}{1.5pt}
\setlength{\dashlinegap}{1.5pt}
\renewcommand{\arraystretch}{1.08}

\resizebox{\textwidth}{!}{%
\begin{tabular}{l|cccc:c|cccc:c|cccc:c|cccc:c}
\hline

\multicolumn{1}{l}{\multirow{2}{*}{Method}}
& \multicolumn{5}{c}{ArtFID$\downarrow$}
& \multicolumn{5}{c}{FID$\downarrow$}
& \multicolumn{5}{c}{LPIPS$\downarrow$}
& \multicolumn{5}{c}{R-FID$\downarrow$} \\

\multicolumn{1}{c}{}
& 2 & 3 & 4 & \multicolumn{1}{c}{5}
& \multicolumn{1}{c}{Avg}
& 2 & 3 & 4 & \multicolumn{1}{c}{5}
& \multicolumn{1}{c}{Avg}
& 2 & 3 & 4 & \multicolumn{1}{c}{5}
& \multicolumn{1}{c}{Avg}
& 2 & 3 & 4 & \multicolumn{1}{c}{5}
& \multicolumn{1}{c}{Avg} \\
\hline

\multicolumn{1}{l|}{\textbf{(G-S)}}
& \multicolumn{4}{c:}{}
& \multicolumn{1}{c|}{}
& \multicolumn{4}{c:}{}
& \multicolumn{1}{c|}{}
& \multicolumn{4}{c:}{}
& \multicolumn{1}{c|}{}
& \multicolumn{4}{c:}{}
& \multicolumn{1}{c}{} \\

$Z^*$
& 17.215 & 17.009 & 17.253 & 16.926 & 17.101
& 10.865 & 10.745 & 10.876 & 10.664 & 10.787
& \underline{0.451} & \underline{0.448} & \underline{0.453}
& \underline{0.451} & \underline{0.451}
& 13.153 & 16.693 & 21.566 & 23.062 & 18.618 \\

StyleID
& 16.635 & 16.819 & 16.995 & 16.927 & 16.844
& 10.323 & 10.451 & 10.643 & 10.383 & 10.450
& 0.469 & 0.469 & 0.460 & 0.487 & 0.471
& 13.685 & 17.956 & 25.562 & 26.523 & 20.931 \\

CoCoDiff
& 16.870 & 17.217 & 17.345 & 17.308 & 17.185
& 10.400 & 10.633 & 10.555 & 10.455 & 10.511
& 0.480 & 0.480 & 0.501 & 0.511 & 0.493
& 13.680 & 18.219 & 26.322 & 27.152 & 21.343 \\

\hline

\multicolumn{1}{l|}{\textbf{(R-S)}}
& \multicolumn{4}{c:}{}
& \multicolumn{1}{c|}{}
& \multicolumn{4}{c:}{}
& \multicolumn{1}{c|}{}
& \multicolumn{4}{c:}{}
& \multicolumn{1}{c|}{}
& \multicolumn{4}{c:}{}
& \multicolumn{1}{c}{} \\

Any-to-Any
& \underline{15.664} & \underline{16.086} & \underline{16.311}
& \underline{16.238} & \underline{16.075}
& \underline{9.090} & \underline{9.379} & \underline{9.498}
& \underline{9.418} & \underline{9.346}
& 0.552 & 0.550 & 0.554 & 0.559 & 0.554
& \textbf{10.691} & 16.390 & \underline{19.751}
& \textbf{20.826} & \underline{16.914} \\

BPC
& 25.631 & 26.015 & 26.311 & 25.384 & 25.835
& 16.332 & 16.726 & 17.057 & 15.346 & 16.365
& 0.479 & 0.468 & 0.457 & 0.553 & 0.489
& 20.614 & 24.709 & 27.718 & 34.020 & 26.765 \\

\hline

\multicolumn{1}{l|}{\textbf{(R-M)}}
& \multicolumn{4}{c:}{}
& \multicolumn{1}{c|}{}
& \multicolumn{4}{c:}{}
& \multicolumn{1}{c|}{}
& \multicolumn{4}{c:}{}
& \multicolumn{1}{c|}{}
& \multicolumn{4}{c:}{}
& \multicolumn{1}{c}{} \\

StyleMixer
& 16.332 & 16.692 & 16.978 & 16.903 & 16.726
& 9.379 & 9.611 & 9.828 & 9.742 & 9.640
& 0.574 & 0.573 & 0.568 & 0.574 & 0.572
& 11.699 & \underline{16.054} & 20.068 & 20.993 & 17.203 \\

StyleGallery
& 18.111 & 19.966 & 20.962 & 20.588 & 19.907
& 11.748 & 13.539 & 14.474 & 14.402 & 13.541
& \textbf{0.421} & \textbf{0.373} & \textbf{0.355}
& \textbf{0.337} & \textbf{0.371}
& 17.658 & 24.866 & 36.256 & 39.409 & 29.547 \\

{\MAST} \textbf{(Ours)}
& \textbf{14.780} & \textbf{14.802} & \textbf{15.014}
& \textbf{15.117} & \textbf{14.928}
& \textbf{8.656} & \textbf{8.700} & \textbf{8.905}
& \textbf{8.929} & \textbf{8.798}
& 0.531 & 0.526 & 0.516 & 0.522 & 0.524
& \underline{11.260} & \textbf{14.219} & \textbf{18.797}
& \underline{20.883} & \textbf{16.290} \\

\hline
\end{tabular}%
}
\caption{
Quantitative comparison across two to five reference styles.
Methods are grouped into Global-Single ({\bf G-S}), Regional-Single
({\bf R-S}), and Regional-Multi ({\bf R-M}) style transfer.
For methods that composite single-style outputs, the best of hard masking
and three soft-blending schemes is reported, selected per setting by ArtFID.
Avg denotes the arithmetic mean over the four style counts.
}
\label{tab:comparison_multiple_styles}
\end{table*}

\begin{figure*}[t]
  \includegraphics[width=\linewidth]{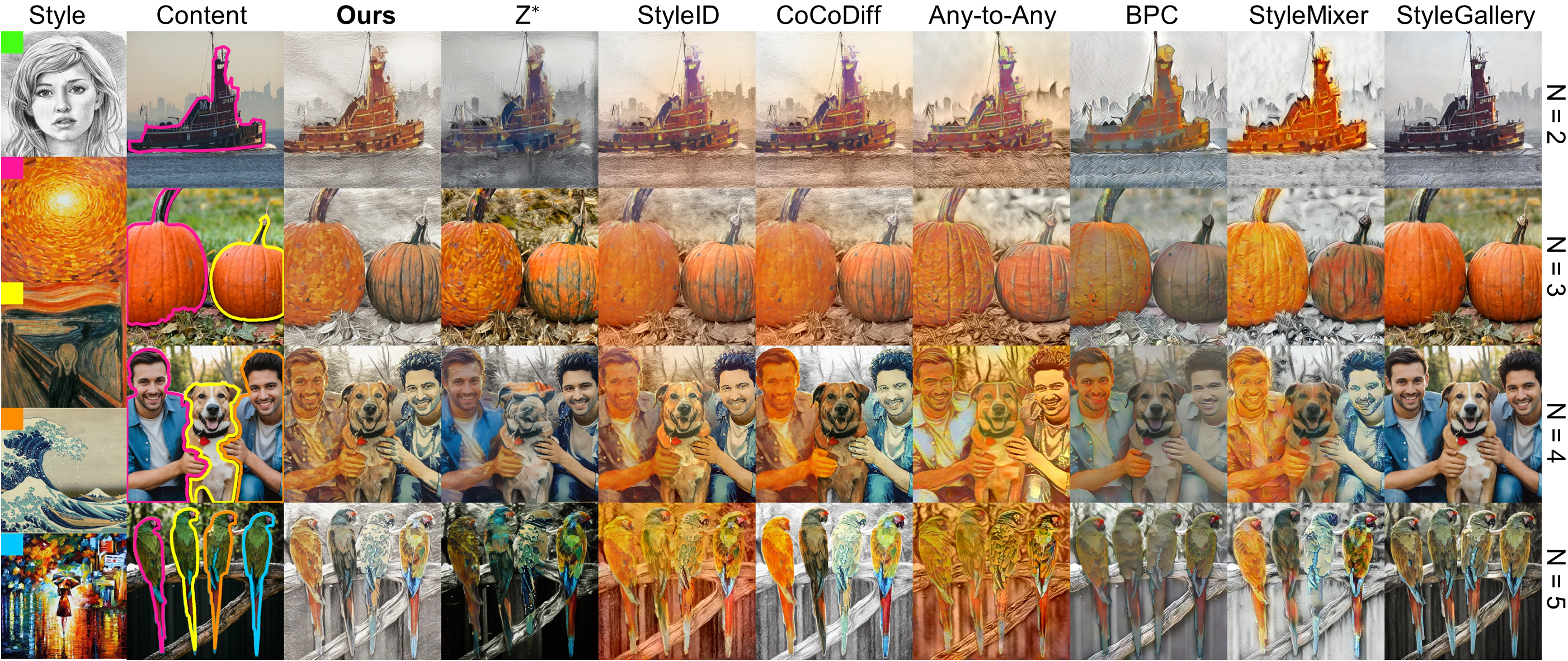}
    \captionof{figure}{
    Qualitative comparison across style settings, illustrating baseline failure modes and {\MAST}'s coherent multi-style results.
    Rows 1--4 correspond to $N = 2, 3, 4, 5$, respectively.
    Columns 4--6, 7--8, and 9--10 show the \textbf{G-S}, \textbf{R-S}, and \textbf{R-M} baselines, respectively.
    Each G-S and R-S baseline uses its best-performing compositing strategy.
    See \cref{fig:teaser} for the mask notation.
    }
  \label{fig:exp_base}
  \vspace{-0.2cm}
\end{figure*}
\section{Experiments}
\label{sec:results}

\subsection{Experimental settings} \label{sec:exp_setting}

\noindent {\bf Datasets.} We use $20$ content images from
MS-COCO~\cite{coco} and $360$ style images from
WikiArt~\cite{wikiart}. For each $N\in\{2,3,4,5\}$, the spatial
partition comprises the $N-1$ largest COCO object regions by pixel area
and one background region, giving $N$ regions in total; FastSAM~\cite{zhao2023fast}
fills otherwise unannotated areas. Grouping the styles into $N$-tuples
and pairing each tuple with all content images yields $3{,}600$,
$2{,}400$, $1{,}800$, and $1{,}440$ samples, respectively.

\noindent {\bf Evaluation metrics.} We evaluate content preservation and style fidelity using FID~\cite{heusel2017gans}, LPIPS~\cite{zhang2018perceptual}, their combined metric ArtFID~\cite{wright2022artfid}, and Region-FID (R-FID), with ArtFID serving as the primary metric because of its correlation with human perception: $\text{ArtFID}=(1+\text{LPIPS})\cdot(1+\text{FID})$. FID measures distributional similarity to the style images, whereas LPIPS measures perceptual similarity to the content image. Because FID cannot verify whether each region matches its assigned style and direct regional-to-global comparison introduces area-dependent bias, we propose a support-matched Monte Carlo metric that resamples style features to match the number of masked-region cells: $\text{R-FID}=\frac{1}{R}\sum_{r=1}^{R}\operatorname{FID}(\mathbf{\Omega}^{o},\mathbf{\Omega}^{s,(r)})$, where $\mathbf{\Omega}^{o}$ and $\mathbf{\Omega}^{s,(r)}$ denote regional output features and equally supported samples from the assigned styles, respectively; see Appendix \cref{app:region_fid} for details.

\noindent {\bf Implementation details.} Experiments use Stable Diffusion v1.4~\cite{stablediffusion} with 50-step DDIM sampling, $\lambda=0.2$, $\pi^\star=0.9$, $r=0.3$, and empty text prompts; following training-free approaches~\cite{StyleID, zstar}, LAMA and STS operate within self-attention modules of U-Net decoder output blocks 7--12 at all denoising timesteps, where high-level semantic interactions are prominent, while DDI restores high-frequency content details in corresponding residual blocks.

\noindent {\bf Baselines.} Using official implementations, we compare {\MAST} with {\bf G-S} ($Z^\ast$~\cite{zstar}, StyleID~\cite{StyleID}, CoCoDiff~\cite{CoCoDiff}), {\bf R-S} (Any-to-Any~\cite{Any-to-any}, BPC~\cite{BPC}), and {\bf R-M} (StyleMixer~\cite{StyleMixer}, StyleGallery~\cite{StyleGallery}) baselines; {\bf G-S} and {\bf R-S} generate one output per style--region pair and apply hard-mask, Alpha~\cite{alpha}, Poisson~\cite{poisson}, or Laplacian-pyramid~\cite{laplacian} compositing, with the best strategy per (method, $N$) selected by dataset-level ArtFID; see
Appendix \cref{appendix:blending} for compositing details and
strategy-wise results.
{\bf R-M} methods use the same inputs, masks, and assignments as {\MAST} without post-hoc composition. 
Throughout \cref{sec:results}, bold and underline denote the best and second-best results, unless otherwise noted.

\subsection{Comparison with baselines} \label{sec:comp_baseline}

\noindent {\bf Quantitative comparison.} As shown in \cref{tab:comparison_multiple_styles}, {\MAST} achieves the most favorable balance between style fidelity and content preservation, attaining the lowest average ArtFID ($14.928$, $7.1\%$ lower than the next-best baseline) and FID ($8.798$) across two to five styles and outperforming Any-to-Any ({\bf R-S} baseline), the strongest competing method on these metrics. Its stability is reflected in the narrow ranges of ArtFID ($14.780$--$15.117$) and FID ($8.656$--$8.929$) as $N$ increases. {\MAST} also achieves the best average R-FID ($16.290$, versus $16.914$ for Any-to-Any and $17.203$ for StyleMixer), indicating faithful region-wise style transfer rather than only improved global appearance.

Although {\MAST} does not attain the lowest LPIPS, this metric alone does not capture regional style fidelity. For example, StyleGallery's LPIPS decreases from $0.421$ to $0.337$ as $N$ increases, while its ArtFID deteriorates from $18.111$ to $20.588$ and its FID from $11.748$ to $14.402$, demonstrating that high content similarity alone is insufficient to assess whether styles have been faithfully transferred.

\begin{figure}[t]
  \includegraphics[width=1.0\linewidth]{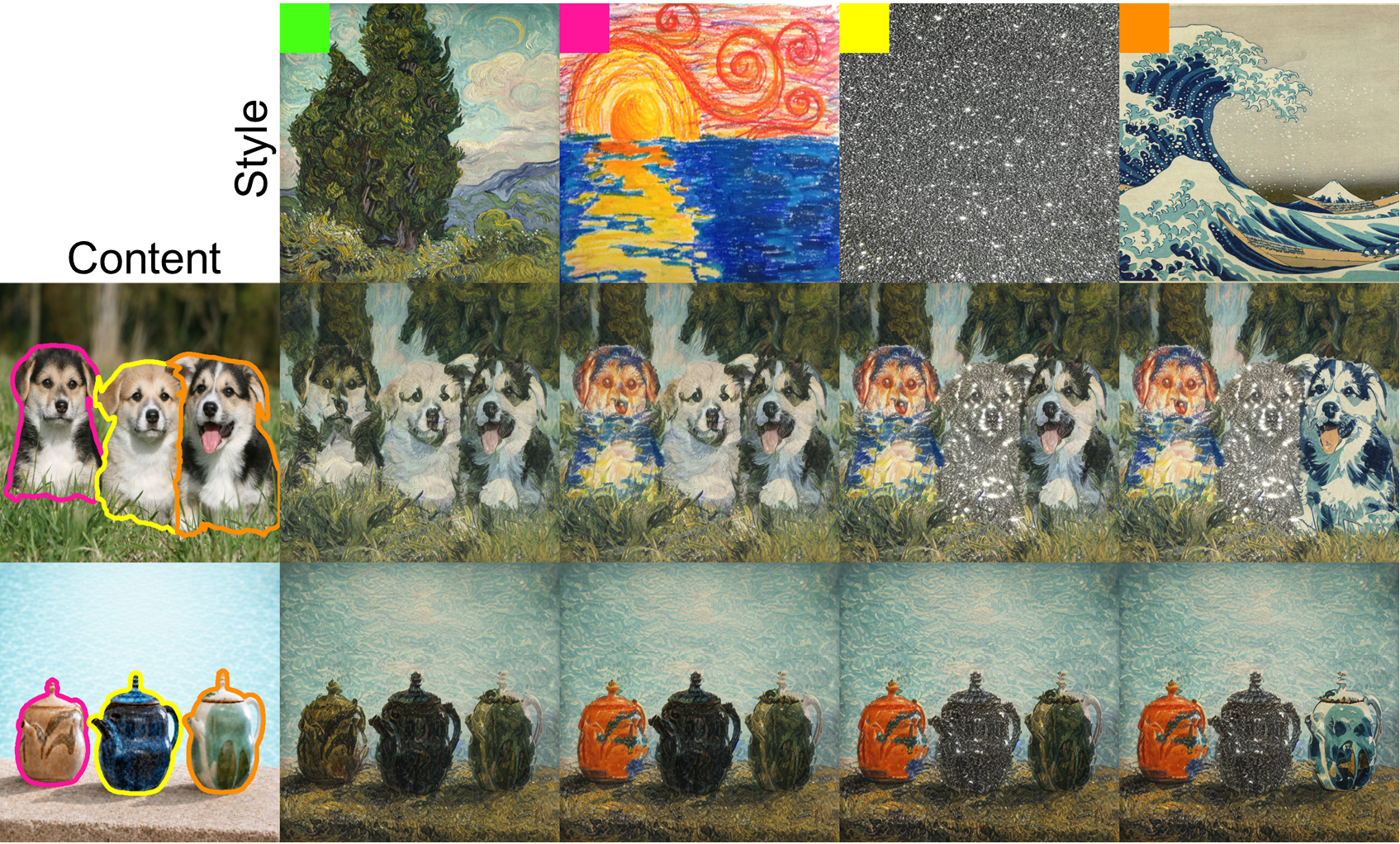}
  \caption{Progressive multi-style transfer under a fixed four-style mask. Styles are sequentially added from left to right.}
  \label{fig:sequential_style}
  \vspace{-0.2cm}
\end{figure}

\noindent {\bf Qualitative comparison.} \cref{fig:exp_base} compares {\MAST} with style transfer baselines and shows three failures. First, $Z^\ast$ and BPC fail to transfer designated styles to target regions in row 1, while StyleGallery produces only weak stylistic changes. Second, single-style methods such as StyleID and CoCoDiff blur fine details, including facial features in row 3, while StyleMixer introduces structural distortions; third, Any-to-Any produces overly accentuated textures and artifacts in row 4. For baselines requiring post-hoc composition, even their best strategies leave visible discontinuities between adjacent regions, showing that compositing alone is insufficient for seamless multi-style integration. 

In contrast, {\MAST} addresses these limitations within a unified attention-control framework: LAMA enforces region-specific style allocation, STS adaptively restores joint-attention selectivity, and DDI preserves fine content details, while its single-path formulation jointly models adjacent regions to produce smoother boundaries than post-hoc compositing. 
To further examine multi-style robustness, \cref{fig:sequential_style} progressively adds one to four styles under a fixed four-style mask; previously applied styles remain stable and newly introduced styles integrate naturally, demonstrating minimal interference and coherent composition. A user study further assesses perceptual quality, with detailed results reported in Appendix \cref{app:user_study}.

\begin{table}[t]
\centering
\setlength{\tabcolsep}{2.5pt}
\renewcommand{\arraystretch}{1.05}
\resizebox{\columnwidth}{!}{%
\begin{tabular}{l|cccc|cccc|cccc}
\toprule
\multirow{2}{*}{Method}
& \multicolumn{4}{c|}{ArtFID$\downarrow$}
& \multicolumn{4}{c|}{FID$\downarrow$}
& \multicolumn{4}{c}{LPIPS$\downarrow$} \\
& 2 & 3 & 4 & 5
& 2 & 3 & 4 & 5
& 2 & 3 & 4 & 5 \\
\midrule

StyleMixer
& 16.461 & 17.025 & 17.661 & 17.665
& 9.646 & 10.087 & 10.571 & 10.595
& 0.546 & 0.536 & 0.526 & 0.524 \\

StyleGallery
& \underline{15.033} & \underline{15.234}
& \underline{15.359} & \underline{15.321}
& \underline{9.240} & \underline{9.445}
& \underline{9.554} & \underline{9.540}
& \textbf{0.468} & \textbf{0.458}
& \textbf{0.455} & \textbf{0.454} \\

{\MAST}
& \textbf{14.780} & \textbf{14.802}
& \textbf{15.014} & \textbf{15.117}
& \textbf{8.656} & \textbf{8.700}
& \textbf{8.905} & \textbf{8.929}
& \underline{0.531} & \underline{0.526}
& \underline{0.516} & \underline{0.522} \\

\bottomrule
\end{tabular}%
}
\caption{Quantitative comparison with automatic semantic-matching methods.}
\label{tab:semantic_comparison}
\end{table}

\noindent {\bf Comparison with automatic semantic-matching methods.} 
Beyond the controlled primary comparison, we evaluate StyleMixer and StyleGallery under their native automatic semantic-matching protocols, while {\MAST} uses the user-provided masks from the primary setting. Because the spatial assignments differ, \cref{tab:semantic_comparison} compares the methods under their intended protocols rather than identical spatial partitions. Under these protocols, {\MAST} consistently achieves the lowest ArtFID and FID, while StyleGallery again exhibits an LPIPS--ArtFID divergence, reinforcing that LPIPS alone does not capture regional style fidelity.

\begin{figure}[t]
  \includegraphics[width=1.0\linewidth]{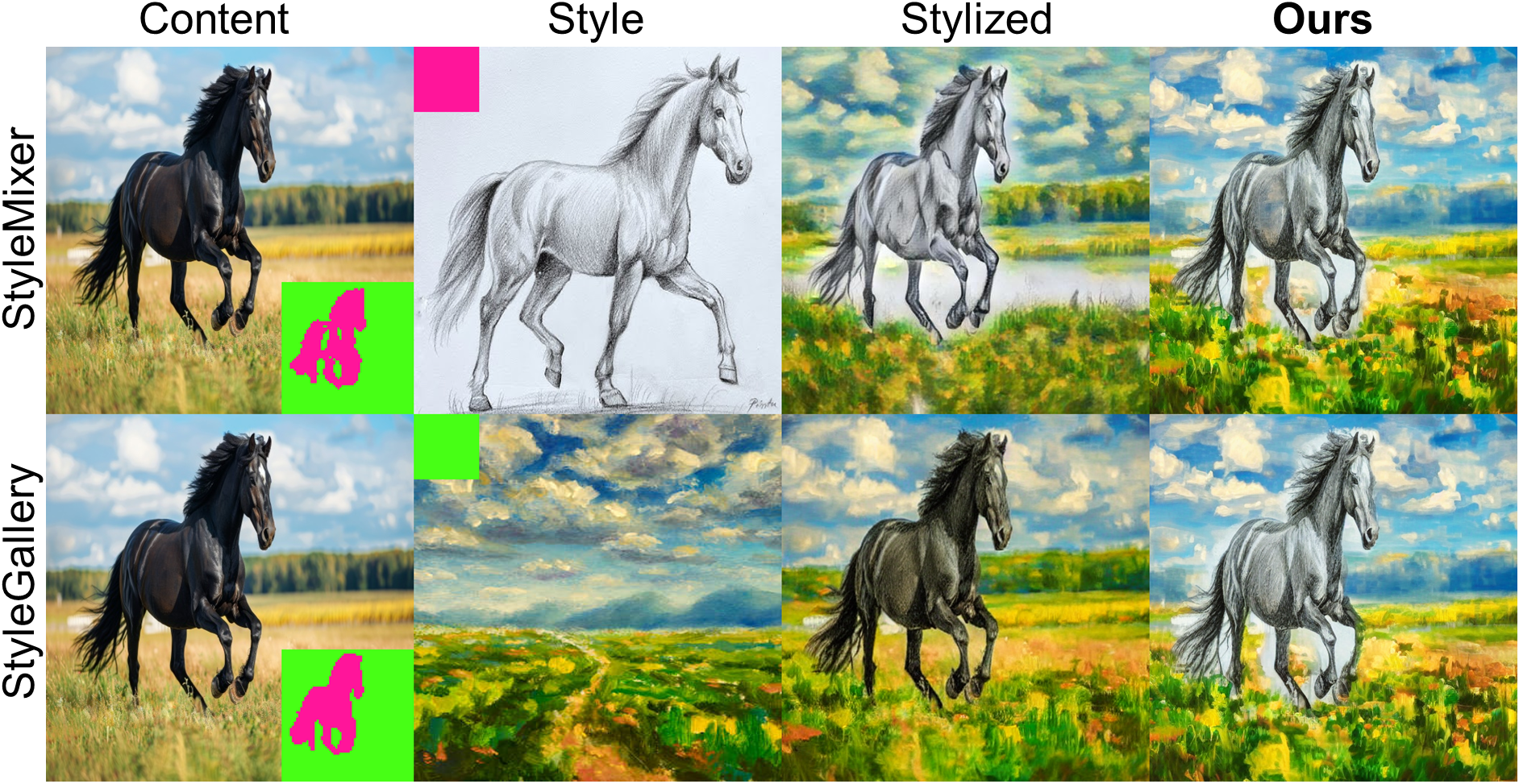}
\captionof{figure}{Comparison with regional-multi baselines under automatic semantic-matching. StyleMixer (top) and StyleGallery (bottom) are shown with their inferred masks (pink and green); {\MAST} uses the same inferred masks for region-wise stylization.
} 
  \label{fig:exp_analysis_sem}
  \vspace{-0.25cm}
\end{figure}

Separately, \cref{fig:exp_analysis_sem} evaluates {\MAST} with the regions inferred by each baseline to align region localization. StyleMixer exhibits style leakage, whereas StyleGallery retains more of the horse's original dark appearance. Given the same inferred regions, {\MAST} produces distinct regional styles with coherent boundaries, demonstrating its applicability beyond the primary user-provided mask protocol.

\subsection{Ablation studies} \label{sec:ablation}

\begin{figure}[t]
\centering
\includegraphics[width=.95\linewidth]{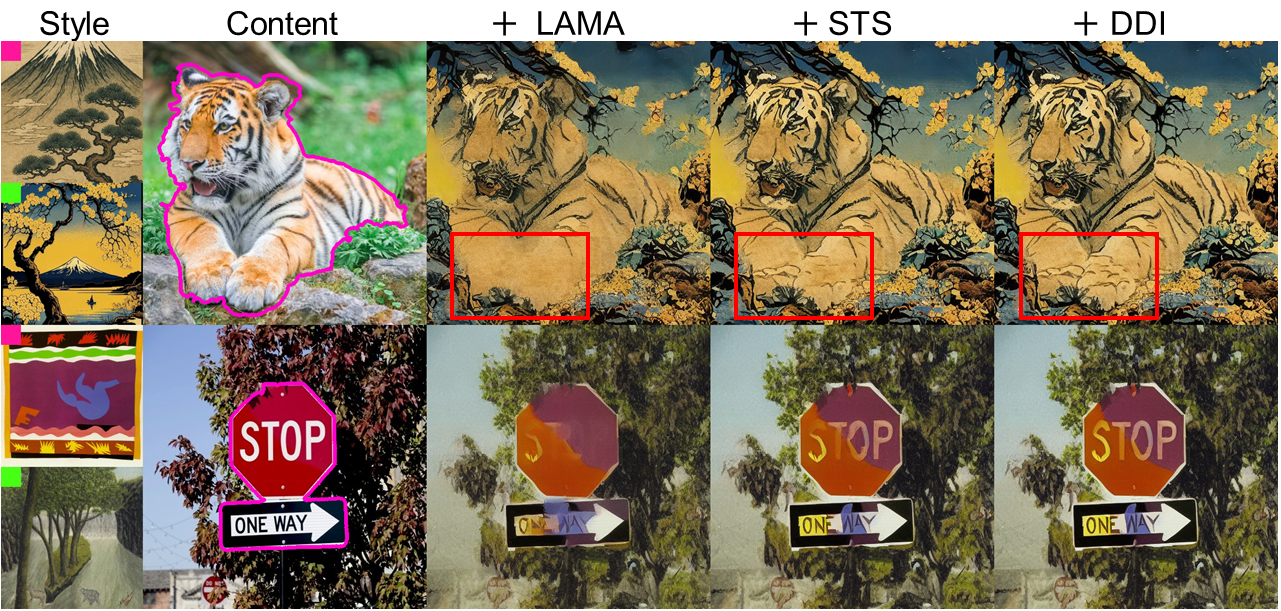}
\caption{
Qualitative ablation of {\MAST} components in the two-style setting. 
}
\label{fig:ablation}
\vspace{-0.3cm}
\end{figure}

\noindent {\bf Overall component analysis.} 
% 위 기존, 아래 예령 수정
We conduct an ablation in the two-style setting (\cref{fig:ablation} and \cref{tb:ablation}), starting from content-anchored query blending and sequentially adding LAMA and STS to address partition-allocation ambiguity and attention-sharpness degradation, respectively, followed by DDI for complementary output-stage detail recovery. 
The base preserves the layout but lacks explicit partition-mass control; its low LPIPS and poor ArtFID, FID, and R-FID therefore indicate content-dominant stylization. 

LAMA substantially improves global and region-wise stylization by enforcing mask-derived query-wise partition masses at the mass-allocation stage; however, the increased style contribution can obscure details such as the tiger-paw pattern in the first row. 
Building on the LAMA-adjusted logits, STS adapts the sharpening factor to the sharpness gap from content self-attention during Softmax normalization, further improving all metrics and yielding clearer, more distinctive style patterns. 
Finally, DDI compensates for high-frequency detail suppression during output-stage feature aggregation through discrepancy-weighted injection, making the ``STOP'' text in the second row clearer. It improves LPIPS with only minor trade-offs in ArtFID ($14.769 \rightarrow 14.780$), FID, and R-FID, and is therefore retained by default to prioritize content-detail preservation.

\begin{table}[t]
\centering

\resizebox{\linewidth}{!}{
\begin{tabular}{ccc|cccc}
\hline
LAMA & STS & DDI
& ArtFID $\downarrow$
& FID $\downarrow$
& LPIPS $\downarrow$
& R-FID $\downarrow$ \\
\hline

& &
& $30.549$
& $24.134$
& $0.216$
& $33.930$ \\

\hdashline

$\checkmark$ & &
& $16.680$
& $9.665$
& $0.564$
& $13.276$ \\

$\checkmark$ & $\checkmark$ &
& $\textbf{14.769}$
& $\textbf{8.495}$
& $\underline{0.555}$
& $\textbf{10.803}$ \\

$\checkmark$ & $\checkmark$ & $\checkmark$
& $\underline{14.780}$
& $\underline{8.656}$
& $\textbf{0.531}$
& $\underline{11.260}$ \\

\hline
\end{tabular}
}
\caption{
Quantitative ablation of {\MAST} components in the two-style setting;
the base setting is excluded from ranking.
}
\label{tb:ablation}
\end{table}

\noindent {\bf LAMA analysis.}
\label{sec:lama_ana}
We compare the mask-derived target distribution in
Eq.~\eqref{eq:lama_target} with the partition masses realized by the shared Softmax before and after LAMA. For an interior query assigned to style $i$, the reported masses satisfy 
$\tilde{\pi}_i+\tilde{\pi}_c+\sum_{k\neq i}\tilde{\pi}_k\simeq1$,
where the last term denotes cross-style leakage, i.e., attention mass assigned to unrelated style partitions. 
We additionally report total variation (TV) for absolute mass displacement from the target, and Jensen--Shannon divergence (JSD) for distribution-level discrepancy. Style, content, and leakage are measured over unambiguous mask interiors.

As shown in \cref{tab:lama_allocation_analysis}, shared attention assigns only $0.242$ mass to the intended style, while $0.582$ remains on content and $0.175$ leaks to the unrelated style, yielding TV $=0.656$
and JSD $=0.276$. LAMA instead realizes the prescribed
$0.900/0.100$ style--content allocation, while leakage, TV, and JSD decrease to numerical precision. This directly confirms the partition-allocation ambiguity identified in \cref{sub:analysis} and shows that LAMA resolves it by enforcing the mask-derived partition masses.

\begin{figure}[t]
\centering

\begin{subfigure}[t]{0.48\linewidth}
    \centering
    \includegraphics[width=\linewidth]{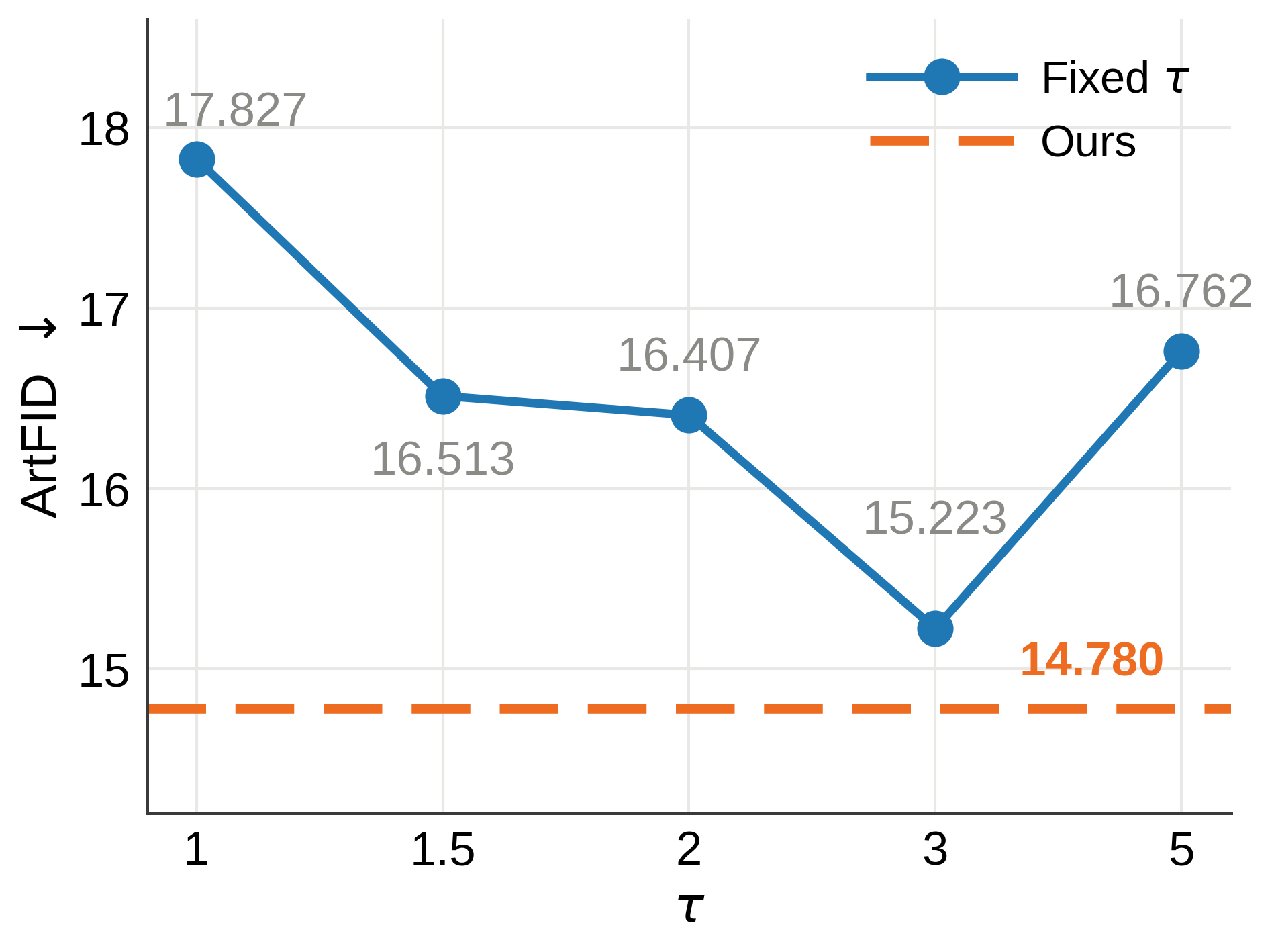}
    \caption{ArtFID: fixed $\tau$ vs.\ STS}
    \label{fig:sts_artfid}
\end{subfigure}
\hfill
\begin{subfigure}[t]{0.48\linewidth}
    \centering
    \includegraphics[width=\linewidth]{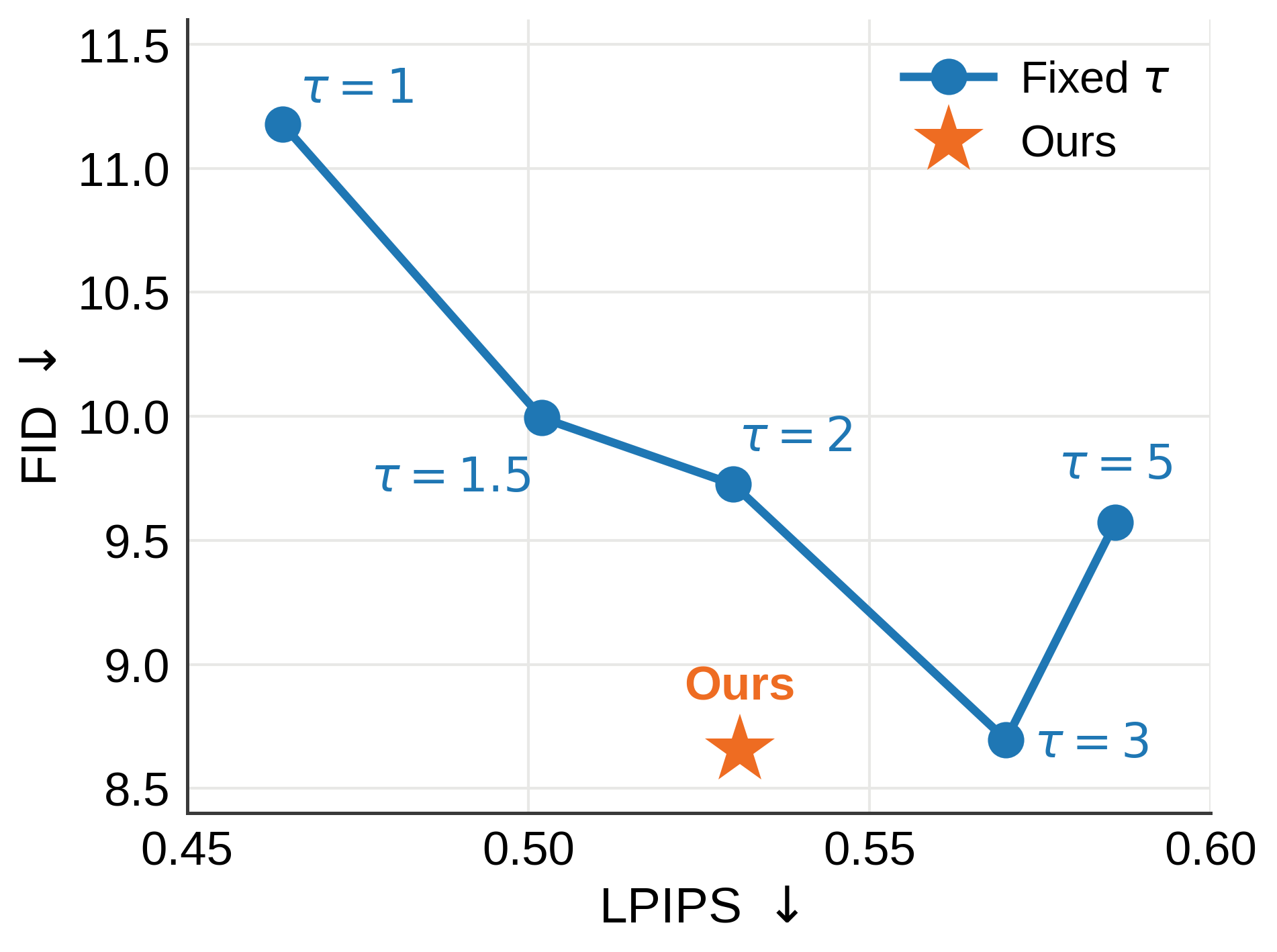}
    \caption{FID--LPIPS trade-off}
    \label{fig:sts_tradeoff}
\end{subfigure}

\caption{
Comparison of fixed and adaptive sharpening in the two-style setting.
}
\label{fig:sts_analysis}
\vspace{-0.2cm}
\end{figure}

\noindent {\bf STS analysis.}
STS achieves a more favorable balance between stylization quality and content
preservation than fixed sharpening. To verify this, we compare STS
with fixed factors in the two-style setting, including $\tau=1.5$ adopted in
StyleID~\cite{StyleID}, while keeping other components unchanged.
As shown in \cref{fig:sts_artfid}, increasing the fixed factor initially
improves ArtFID by correcting insufficiently sharp attention, but excessive
scaling reverses this gain; the dashed line indicates the result of adaptive
STS. \cref{fig:sts_tradeoff} further shows that stronger fixed
sharpening improves FID at the cost of LPIPS before excessive scaling
eventually degrades both.

These results demonstrate that no single sharpening strength is consistently
effective. STS instead adapts $\tau$ to the sharpness gap from content
self-attention, providing the required correction for the current attention
distribution. This supports content-reference-guided adaptation over a
globally fixed factor.

\begin{table}[t]
\centering

\resizebox{\linewidth}{!}{%
\begin{tabular}{l|ccccc}
\hline
Method
& Style
& Content
& Leakage $\downarrow$
& TV $\downarrow$
& JSD $\downarrow$ \\
\hline
Shared Attn.
& 0.242
& 0.582
& 0.175
& 0.656
& 0.276 \\
+ LAMA
& $\mathbf{0.900}$
& $\mathbf{0.100}$
& $\mathbf{1.0 \times 10^{-6}}$
& $\mathbf{9.1 \times 10^{-8}}$
& $\mathbf{1.6 \times 10^{-8}}$ \\
Target
& 0.900
& 0.100
& 0
& 0
& 0 \\
\hline
\end{tabular}%
}
\caption{
Partition-allocation ambiguity in the two-style setting over $200$ images, resolved by LAMA. Leakage, TV, and JSD reach numerical precision ($< 10^{-5}$) under LAMA.
}
\label{tab:lama_allocation_analysis}
\end{table}

\section{Conclusion}
\label{sec:conclusion}

We presented {\MAST}, a training-free framework for regional-multi style transfer built on a shared attention analysis: masks alone cannot resolve partition-level competition, joint normalization reduces selectivity, and feature aggregation suppresses fine content details. {\MAST} addresses these through LAMA (mass allocation), STS (attention sharpness), and DDI (detail recovery). Across two- to five-style settings, {\MAST} achieves the best average ArtFID, FID, and R-FID among all baselines, demonstrating the effectiveness of controlling content-style interactions within shared attention. 
{\MAST} currently assumes user-provided masks, and our evaluation is limited to five styles because increasing the number of partitions progressively reduces the spatial support available to each style. Extending the framework to denser regional assignments while reducing the required mask specification remains future work.

{
    \small
    \bibliographystyle{ieeenat_fullname}
    \bibliography{main}

@String(TOG= {ACM Trans. Graph.})

@String(AAAI = {AAAI})

@String(TOG   = {ACM TOG})

@inproceedings{stablediffusion,
  title={High-resolution image synthesis with latent diffusion models},
  author={Rombach, Robin and Blattmann, Andreas and Lorenz, Dominik and Esser, Patrick and Ommer, Bj{\"o}rn},
  booktitle={Proceedings of the IEEE/CVF Conference on Computer Vision and Pattern Recognition},
  pages={10684--10695},
  year={2022}
}

@inproceedings{heusel2017gans,
  title={{GANs} Trained by a Two Time-Scale Update Rule Converge to a Local Nash Equilibrium},
  author={Heusel, Martin and Ramsauer, Hubert and Unterthiner, Thomas and Nessler, Bernhard and Hochreiter, Sepp},
  booktitle={Advances in Neural Information Processing Systems},
  volume={30},
  year={2017}
}

@inproceedings{zhang2018perceptual,
  title={The Unreasonable Effectiveness of Deep Features as a Perceptual Metric},
  author={Zhang, Richard and Isola, Phillip and Efros, Alexei A. and Shechtman, Eli and Wang, Oliver},
  booktitle={Proceedings of the IEEE/CVF Conference on Computer Vision and Pattern Recognition},
  pages={586--595},
  year={2018}
}

@inproceedings{wright2022artfid,
  title={{ArtFID}: Quantitative Evaluation of Neural Style Transfer},
  author={Wright, Matthias and Ommer, Bj{\"o}rn},
  booktitle={DAGM German Conference on Pattern Recognition},
  pages={560--576},
  year={2022},
  publisher={Springer}
}

@article{zhao2023fast,
  title={Fast segment anything},
  author={Zhao, Xu and Ding, Wenchao and An, Yongqi and Du, Yinglong and Yu, Tao and Li, Min and Tang, Ming and Wang, Jinqiao},
  journal={arXiv preprint arXiv:2306.12156},
  year={2023}
}

@inproceedings{ddim,
  title={Denoising Diffusion Implicit Models},
  author={Song, Jiaming and Meng, Chenlin and Ermon, Stefano},
  booktitle={International Conference on Learning Representations},
  year={2021},
  url={https://openreview.net/forum?id=St1giarCHLP}
}

@article{Consislora,
  title={Consislora: Enhancing content and style consistency for lora-based style transfer},
  author={Chen, Bolin and Zhao, Baoquan and Xie, Haoran and Cai, Yi and Li, Qing and Mao, Xudong},
  journal={arXiv preprint arXiv:2503.10614},
  year={2025}
}

@inproceedings{Stylediffusion,
  title={Stylediffusion: Controllable disentangled style transfer via diffusion models},
  author={Wang, Zhizhong and Zhao, Lei and Xing, Wei},
  booktitle={Proceedings of the IEEE/CVF International Conference on Computer Vision},
  pages={7677--7689},
  year={2023}
}

@inproceedings{clip,
  title={Learning transferable visual models from natural language supervision},
  author={Radford, Alec and Kim, Jong Wook and Hallacy, Chris and Ramesh, Aditya and Goh, Gabriel and Agarwal, Sandhini and Sastry, Girish and Askell, Amanda and Mishkin, Pamela and Clark, Jack and others},
  booktitle={International conference on machine learning},
  pages={8748--8763},
  year={2021},
  organization={PmLR}
}

@article{DiffStyle,
  title={Training-Free Style Transfer Emerges from h-Space in Diffusion Models},
  author={Jeong, Jaeseok and Kwon, Mingi and Uh, Youngjung},
  journal={arXiv preprint arXiv:2303.15403},
  year={2023}
}

@article{TFLDST,
  title={A Training-Free Latent Diffusion Style Transfer Method},
  author={Xiang, Zhengtao and Wan, Xing and Xu, Libo and Yu, Xin and Mao, Yuhan},
  journal={Information},
  volume={15},
  number={10},
  pages={588},
  year={2024},
  publisher={MDPI}
}

@article{AttenST,
  title={AttenST: A Training-Free Attention-Driven Style Transfer Framework with Pre-Trained Diffusion Models},
  author={Huang, Bo and Xu, Wenlun and Han, Qizhuo and Jing, Haodong and Li, Ying},
  journal={arXiv preprint arXiv:2503.07307},
  year={2025}
}

@inproceedings{FineGrainedST,
  title={Fine-grained image style transfer with visual transformers},
  author={Wang, Jianbo and Yang, Huan and Fu, Jianlong and Yamasaki, Toshihiko and Guo, Baining},
  booktitle={Proceedings of the Asian Conference on Computer Vision},
  pages={841--857},
  year={2022}
}

@inproceedings{coco,
  title={Microsoft {COCO}: Common Objects in Context},
  author={Lin, Tsung-Yi and Maire, Michael and Belongie, Serge and Hays, James and Perona, Pietro and Ramanan, Deva and Doll{\'a}r, Piotr and Zitnick, C. Lawrence},
  booktitle={European Conference on Computer Vision},
  pages={740--755},
  year={2014},
  doi={10.1007/978-3-319-10602-1_48}
}

@inproceedings{diffusest,
  title={Diffusest: Unleashing the capability of the diffusion model for style transfer},
  author={Hu, Ying and Zhuang, Chenyi and Gao, Pan},
  booktitle={Proceedings of the 6th ACM International Conference on Multimedia in Asia},
  pages={1--1},
  year={2024}
}

@inproceedings{stylessp,
  title={Stylessp: Sampling startpoint enhancement for training-free diffusion-based method for style transfer},
  author={Xu, Ruojun and Xi, Weijie and Wang, XiaoDi and Mao, Yongbo and Cheng, Zach},
  booktitle={Proceedings of the Computer Vision and Pattern Recognition Conference},
  pages={18260--18269},
  year={2025}
}

@inproceedings{velivckovic2024softmax,
  title={Softmax Is Not Enough (for Sharp Size Generalisation)},
  author={Veli{\v{c}}kovi{\'c}, Petar and Perivolaropoulos, Christos and Barbero, Federico and Pascanu, Razvan},
  booktitle={Proceedings of the 42nd International Conference on Machine Learning},
  series={Proceedings of Machine Learning Research},
  volume={267},
  pages={61190--61211},
  year={2025},
  publisher={PMLR},
  url={https://proceedings.mlr.press/v267/velickovic25a.html}
}

@misc{wikiart,
  author = {{WikiArt}},
  title = {WikiArt: Visual Art Encyclopedia},
  howpublished = {\url{https://www.wikiart.org/}},
  year = {2026},
  note = {Accessed: 2026-03-22}
}

@article{res-based2,
  title={Training-free style transfer via content-style image inversion},
  author={Lei, Songlin and Yang, Qiuxia and Yang, Ke and Zhao, Zhengpeng and Pu, Yuanyuan},
  journal={Computers \& Graphics},
  pages={104352},
  year={2025},
  publisher={Elsevier}
}

@article{learned,
  title={A learned representation for artistic style},
  author={Dumoulin, Vincent and Shlens, Jonathon and Kudlur, Manjunath},
  journal={arXiv preprint arXiv:1610.07629},
  year={2016}
}

@inproceedings{StyCNN,
  title={Image style transfer using convolutional neural networks},
  author={Gatys, Leon A and Ecker, Alexander S and Bethge, Matthias},
  booktitle={Proceedings of the IEEE conference on computer vision and pattern recognition},
  pages={2414--2423},
  year={2016}
}

@inproceedings{Adain,
  title={Arbitrary style transfer in real-time with adaptive instance normalization},
  author={Huang, Xun and Belongie, Serge},
  booktitle={Proceedings of the IEEE International Conference on Computer Vision},
  pages={1501--1510},
  year={2017}
}

@article{universal,
  title={Universal style transfer via feature transforms},
  author={Li, Yijun and Fang, Chen and Yang, Jimei and Wang, Zhaowen and Lu, Xin and Yang, Ming-Hsuan},
  journal={Advances in Neural Information Processing Systems},
  volume={30},
  year={2017}
}

@InProceedings{StyTr2,
    author    = {Deng, Yingying and Tang, Fan and Dong, Weiming and Ma, Chongyang and Pan, Xingjia and Wang, Lei and Xu, Changsheng},
    title     = {StyTr2: Image Style Transfer With Transformers},
    booktitle = {Proceedings of the IEEE/CVF Conference on Computer Vision and Pattern Recognition},
    month     = {June},
    year      = {2022},
    pages     = {11326-11336}
}

@inproceedings{InST,
  title={Inversion-based style transfer with diffusion models},
  author={Zhang, Yuxin and Huang, Nisha and Tang, Fan and Huang, Haibin and Ma, Chongyang and Dong, Weiming and Xu, Changsheng},
  booktitle={2023 IEEE/CVF Conference on Computer Vision and Pattern Recognition},
  pages={10146--10156},
  year={2023},
  organization={IEEE}
}

@inproceedings{zstar,
  title={Z*: Zero-shot style transfer via attention reweighting},
  author={Deng, Yingying and He, Xiangyu and Tang, Fan and Dong, Weiming},
  booktitle={Proceedings of the IEEE/CVF Conference on Computer Vision and Pattern Recognition},
  pages={6934--6944},
  year={2024}
}

@InProceedings{StyleID,
    author    = {Chung, Jiwoo and Hyun, Sangeek and Heo, Jae-Pil},
    title     = {Style Injection in Diffusion: A Training-free Approach for Adapting Large-scale Diffusion Models for Style Transfer},
    booktitle = {Proceedings of the IEEE/CVF Conference on Computer Vision and Pattern Recognition},
    month     = {June},
    year      = {2024},
    pages     = {8795-8805}
}

@article{StyleFM,
  title={{StyleFM}: Frequency Manipulation Empowered by Recursive Attention on Diffusion Models for Arbitrary Style Transfer},
  author={Ma, Yingnan and Liu, Zhenye and Liu, Siying and Basu, Anup},
  journal={Proceedings of the AAAI Conference on Artificial Intelligence},
  volume={40},
  number={10},
  pages={7865--7873},
  year={2026},
  doi={10.1609/aaai.v40i10.37730}
}

@inproceedings{CoCoDiff,
  title={{CoCoDiff}: Correspondence-Consistent Diffusion Model for Fine-grained Style Transfer},
  author={Nie, Wenbo and Li, Zixiang and Tao, Renshuai and Wu, Bin and Wei, Yunchao and Zhao, Yao},
  booktitle={The Fourteenth International Conference on Learning Representations},
  year={2026},
  url={https://openreview.net/forum?id=eRSuwB78LH}
}

@article{wang2024multisource,
  title={Multi-Source Style Transfer via Style Disentanglement Network},
  author={Wang, Quan and Li, Sheng and Wang, Zichi and Zhang, Xinpeng and Feng, Guorui},
  journal={IEEE Transactions on Multimedia},
  volume={26},
  pages={1373--1383},
  year={2024},
  doi={10.1109/TMM.2023.3281087}
}

@article{yu2025multisource,
  title={Multi-Source Training-Free Controllable Style Transfer via Diffusion Models},
  author={Yu, Cuihong and Han, Cheng and Zhang, Chao},
  journal={Symmetry},
  volume={17},
  number={2},
  pages={290},
  year={2025},
  doi={10.3390/sym17020290}
}

@inproceedings{ruta2025leveraging,
  title={Leveraging Diffusion Models for Stylization Using Multiple Style Images},
  author={Ruta, Dan and Djelouah, Abdelaziz and Ortiz, Raphael and Schroers, Christopher},
  booktitle={Proceedings of the IEEE/CVF International Conference on Computer Vision Workshops},
  pages={417--426},
  year={2025}
}

@inproceedings{SCAdapter,
  title={{SCAdapter}: Content-Style Disentanglement for Diffusion Style Transfer},
  author={Trinh, Luan Thanh and Doi, Kenji and Osanai, Atsuki},
  booktitle={Proceedings of the IEEE/CVF Winter Conference on Applications of Computer Vision},
  pages={7312--7321},
  year={2026}
}

@article{kim2024controllable,
  title={Controllable Style Transfer via Test-time Training of Implicit Neural Representation},
  author={Kim, Sunwoo and Min, Youngjo and Jung, Younghun and Kim, Seungryong},
  journal={Pattern Recognition},
  volume={146},
  pages={109988},
  year={2024},
  doi={10.1016/j.patcog.2023.109988}
}

@article{Any-to-any,
  title={Any-to-any style transfer: Making picasso and da vinci collaborate},
  author={Liu, Songhua and Ye, Jingwen and Wang, Xinchao},
  journal={arXiv preprint arXiv:2304.09728},
  year={2023}
}

@inproceedings{chen2026regionroute,
  title={{RegionRoute}: Regional Style Transfer with Diffusion Model},
  author={Chen, Bowen and Zuena, Jake and Bovik, Alan C. and Kothandaraman, Divya},
  booktitle={Proceedings of the IEEE/CVF Conference on Computer Vision and Pattern Recognition},
  year={2026},
  url={https://arxiv.org/abs/2602.19254}
}

@article{StyleMixer,
  title={{Style Mixer}: Semantic-Aware Multi-Style Transfer Network},
  author={Huang, Zixuan and Zhang, Jinghuai and Liao, Jing},
  journal={Computer Graphics Forum},
  volume={38},
  number={7},
  pages={469--480},
  year={2019},
  doi={10.1111/cgf.13853}
}

@article{ye2023multistyle,
  title={Multi-Style Transfer and Fusion of Image's Regions Based on Attention Mechanism and Instance Segmentation},
  author={Ye, Wujian and Liu, Chaojie and Chen, Yuehai and Liu, Yijun and Liu, Chenming and Zhou, Huihui},
  journal={Signal Processing: Image Communication},
  volume={110},
  pages={116871},
  year={2023},
  doi={10.1016/j.image.2022.116871}
}

@article{BPC,
  title={Improving Masked Style Transfer Using Blended Partial Convolution},
  author={Seyed, Seyedhadi and Cansever, Ayberk and Hart, David},
  journal={IEEE Access},
  volume={14},
  pages={64584--64595},
  year={2026},
  doi={10.1109/ACCESS.2026.3687089}
}

@article{li2026efficient,
  title={An Efficient and Harmonized Framework for Balanced Cross-Domain Feature Integration},
  author={Li, Shaoxu and Pan, Ye},
  journal={Proceedings of the AAAI Conference on Artificial Intelligence},
  volume={40},
  number={8},
  pages={6369--6377},
  year={2026},
  doi={10.1609/aaai.v40i8.37564}
}

@inproceedings{StyleGallery,
  title={StyleGallery: Training-free and Semantic-aware Personalized Style Transfer from Arbitrary Image References},
  author={He, Boyu and Ye, Yunfan and Liu, Chang and Wu, Weishang and Liu, Fang and Cai, Zhiping},
  booktitle={Proceedings of the IEEE/CVF Conference on Computer Vision and Pattern Recognition},
  pages={29092--29102},
  year={2026}
}

@inproceedings{alpha,
  title={Compositing digital images},
  author={Porter, Thomas and Duff, Tom},
  booktitle={Proceedings of the 11th Annual Conference on Computer Graphics and Interactive Techniques},
  pages={253--259},
  year={1984}
}

@incollection{poisson,
  title={Poisson image editing},
  author={P{\'e}rez, Patrick and Gangnet, Michel and Blake, Andrew},
  booktitle={Seminal Graphics Papers: Pushing the Boundaries, Volume 2},
  pages={577--582},
  year={2023}
}

@article{laplacian,
  title={A multiresolution spline with application to image mosaics},
  author={Burt, Peter J and Adelson, Edward H},
  journal={ACM Transactions on Graphics (ToG)},
  volume={2},
  number={4},
  pages={217--236},
  year={1983},
  publisher={ACM New York, NY, USA}
}
}

% =========================
% Appendix
% =========================

\clearpage
\appendix

\appendix

\section{Additional Analyses and Results}

\subsection{Sensitivity analysis of content anchoring parameter \texorpdfstring{$\lambda$}{lambda}}

\begin{table}[h]
\centering

\small
\setlength{\tabcolsep}{5pt}
\renewcommand{\arraystretch}{1.08}
\begin{tabular}{c|ccccc}
\hline
$\lambda$ & ArtFID $\downarrow$ & FID $\downarrow$ & LPIPS $\downarrow$ & R-FID $\downarrow$ \\
\hline
0.50 & 16.690 & 10.459 & \bf{0.457} & 14.147 \\
0.25 & 15.103 & 8.960  & \underline{0.516} & 11.735 \\
0.20 & 14.780 & 8.656  & 0.531 & 11.260 \\
0.15 & \underline{14.645} & \underline{8.476}  & 0.546 & \underline{10.828} \\
0.10 & \bf{14.357} & \bf{8.186}  & 0.563 & \bf{10.112} \\
\hline
\end{tabular}
\caption{Effect of the content anchoring parameter $\lambda$ on performance.}
\label{tab:ablation_lambda}
\end{table}

Following StyleID~\cite{StyleID}, we anchor the content-stylized query to the content query, with $\lambda\in[0,1]$ controlling the strength of content anchoring. To investigate the effect of $\lambda$, we perform a sensitivity analysis over different values, as reported in \cref{tab:ablation_lambda}. As $\lambda$ increases, stronger content anchoring improves content preservation, as reflected by lower LPIPS, but suppresses stylistic information and degrades the other metrics. Conversely, smaller values strengthen global and regional style fidelity at the cost of content similarity. Although $\lambda=0.1$ yields the lowest ArtFID in this sweep, its higher LPIPS indicates greater deviation from the content image. Moreover, because LAMA retains a mask-guided content allocation, excessively strong query anchoring is unnecessary. We therefore set $\lambda=0.2$ as a practical operating point that preserves
content similarity while maintaining strong style fidelity.

\subsection{Mathematical derivation and analysis of LAMA}
\label{app:lama}

We provide additional derivations and analyses of Logit-level Attention
Mass Allocation (LAMA) for the general multi-style setting. Consistent
with the notation used in the subsequent STS analysis, we consider a
single attention head and omit the head index. Let
$L_i\in\mathbb{R}^{N_q\times N_k}$ and
$L_c\in\mathbb{R}^{N_q\times N_k}$ denote the query-by-key logit
matrices of style partition $i$ and the content partition, respectively.
Here, $q$ indexes spatial queries and $j$ indexes keys within a
partition. Concatenating the $N$ style partitions and the content
partition therefore produces a joint logit matrix with
$(N+1)N_k$ keys.

\noindent {\bf Raw and target partition allocation.} For a spatial query $q$, we define the partition functions
\begin{equation}
Z_i(q)
=
\sum_{j=1}^{N_k}
\exp\!\bigl(L_i(q,j)\bigr),
\quad
Z_c(q)
=
\sum_{j=1}^{N_k}
\exp\!\bigl(L_c(q,j)\bigr).
\label{eq:app_partition_functions}
\end{equation}
Before LAMA, the shared Softmax normalizes all style and content
partitions jointly:
\begin{equation}
Z_{\mathrm{raw}}(q)
=
Z_c(q)
+
\sum_{k=1}^{N} Z_k(q).
\label{eq:app_raw_normalization}
\end{equation}
The resulting partition masses are
\begin{equation}
\tilde{\pi}_i^{\mathrm{raw}}(q)
=
\frac{Z_i(q)}{Z_{\mathrm{raw}}(q)},
\qquad
\tilde{\pi}_c^{\mathrm{raw}}(q)
=
\frac{Z_c(q)}{Z_{\mathrm{raw}}(q)}.
\label{eq:app_raw_mass}
\end{equation}

These quantities make the partition-allocation ambiguity in
\cref{sub:analysis} explicit. Although the masks specify which
style should be active at query $q$, they do not appear in
\cref{eq:app_raw_mass}. The realized masses are therefore
determined entirely by the relative unnormalized strengths
$Z_i(q)$ and $Z_c(q)$.

LAMA instead converts the spatial masks into the target distribution
\begin{equation}
\pi_i^\star(q)
=
\pi^\star M_i(q),
\qquad
\pi_c^\star(q)
=
1-\sum_{k=1}^{N}\pi_k^\star(q),
\label{eq:app_target_mass}
\end{equation}
which satisfies
\begin{equation}
\sum_{k=1}^{N}\pi_k^\star(q)+\pi_c^\star(q)=1.
\end{equation}
Thus, $\pi^\star$ is a maximum \emph{total} style-mass budget at a
fully assigned query, rather than an independent budget for every
style partition. The mask values determine how this total budget is
distributed spatially and across styles.

\noindent {\bf Closed-form bias and exact realization.}
For $\pi_i^\star(q)>0$ and $\pi_c^\star(q)>0$, LAMA adds a
query-dependent scalar bias $b_i(q)$ to every key logit in style
partition $i$, while leaving the content partition unbiased. Before
LAMA, the style-to-content mass ratio is $Z_i(q)/Z_c(q)$. After the
bias, it becomes
$\exp(b_i(q))Z_i(q)/Z_c(q)$. Matching this ratio to the target gives
\begin{equation}
\begin{aligned}
&\frac{\exp(b_i(q))Z_i(q)}{Z_c(q)}
=
\frac{\pi_i^\star(q)}{\pi_c^\star(q)},\\
b_i(q)
=
&\log\!\left(
\frac{\pi_i^\star(q)}{\pi_c^\star(q)}
\right)
+
\log Z_c(q)
-
\log Z_i(q).
\end{aligned}
\label{eq:app_closed_form_bias}
\end{equation}
The first term specifies the desired style-to-content ratio, while
the remaining terms remove the dependence on the current raw ratio.
Consequently, a partition whose raw mass is too small is shifted
upward, whereas one whose raw mass is too large is shifted downward.

Applying the bias independently at every query yields
\begin{equation}
\begin{aligned}
L_{\mathrm{LAMA}}(q,:)
&=
\bigl[
L_1(q,:)+b_1(q),\ldots,\\
&\qquad
L_N(q,:)+b_N(q),L_c(q,:)
\bigr].
\end{aligned}
\label{eq:app_lama_logits}
\end{equation}

We next verify that this correction realizes the target masses.
Let
\begin{equation}
Z_{\mathrm{LAMA}}(q)
=
Z_c(q)
+
\sum_{k=1}^{N}
\exp(b_k(q))Z_k(q).
\label{eq:app_lama_normalization}
\end{equation}
denote the normalization of the LAMA-adjusted joint logits.
From \cref{eq:app_closed_form_bias},
\begin{equation}
\exp(b_i(q))Z_i(q)
=
\frac{\pi_i^\star(q)}
{\pi_c^\star(q)}
Z_c(q).
\label{eq:app_bias_identity}
\end{equation}
Using
$\sum_{k=1}^{N}\pi_k^\star(q)
=
1-\pi_c^\star(q)$ gives
\begin{equation}
\begin{aligned}
Z_{\mathrm{LAMA}}(q)
&=
Z_c(q)
+
\sum_{k=1}^{N}
\exp(b_k(q))Z_k(q)
\\
&=
Z_c(q)
\left(
1+
\frac{\sum_{k=1}^{N}\pi_k^\star(q)}
{\pi_c^\star(q)}
\right)
=
\frac{Z_c(q)}{\pi_c^\star(q)}.
\end{aligned}
\label{eq:app_exact_normalization}
\end{equation}

Therefore, the realized partition masses after LAMA are
\begin{equation}
\begin{aligned}
\tilde{\pi}_c^{\mathrm{LAMA}}(q)
&=
\frac{Z_c(q)}{Z_{\mathrm{LAMA}}(q)}
=
\pi_c^\star(q),
\\
\tilde{\pi}_i^{\mathrm{LAMA}}(q)
&=
\frac{\exp(b_i(q))Z_i(q)}
{Z_{\mathrm{LAMA}}(q)}
=
\pi_i^\star(q).
\end{aligned}
\label{eq:app_exact_allocation}
\end{equation}

Hence, a single shared Softmax
over $L_{\mathrm{LAMA}}$ exactly realizes the mask-derived target
distribution at every query. Importantly, this statement characterizes
the LAMA stage before the subsequent STS scaling; STS can later modify
the exact partition masses while largely retaining their spatial
allocation pattern, as analyzed in \cref{appendix:sts_interaction}.

The derivation above assumes strictly positive target masses; boundary
cases containing a zero target mass are understood as the corresponding
limit of the positive-mass solution.

\noindent {\bf Preservation of within-partition correspondence.} Controlling partition mass is useful only if it does not disturb the
query--key correspondence already encoded by the attention logits.
Changing the relative logits within a style partition would alter which
reference keys are retrieved for a given query and could therefore
redirect attention toward unrelated style features.

LAMA avoids this by adding the same scalar $b_i(q)$ to every key in
style partition $i$. Consequently,
\begin{equation}
\operatorname{Softmax}
\bigl(L_i(q,:)+b_i(q)\bigr)
=
\operatorname{Softmax}
\bigl(L_i(q,:)\bigr).
\label{eq:app_intra_partition_invariance}
\end{equation}
Thus, if
\begin{equation}
p_i(j\mid q)
=
\operatorname{Softmax}\bigl(L_i(q,:)\bigr)_j
\end{equation}
denotes the original within-partition selection, the LAMA probability
factorizes as
\begin{equation}
p_{\mathrm{LAMA}}(i,j\mid q)
=
\underbrace{\pi_i^\star(q)}_{\text{partition allocation}}
\underbrace{p_i(j\mid q)}_{\text{within-partition selection}}.
\label{eq:app_probability_factorization}
\end{equation}
Let $V_i\in\mathbb{R}^{N_k\times d_v}$ denote the value matrix of
style partition $i$.
Accordingly, the style-value contribution from partition $i$ becomes
\begin{equation}
\sum_j p_{\mathrm{LAMA}}(i,j\mid q)V_i(j,:)
=
\pi_i^\star(q)
\sum_j p_i(j\mid q)V_i(j,:).
\label{eq:app_partition_value_contribution}
\end{equation}
LAMA therefore controls \emph{how strongly} each style partition
contributes without changing \emph{which reference features} are
selected within that partition. This preserves the correspondence
structure learned by the pretrained attention while modifying only
the competition among style and content partitions.

\noindent {\bf Quantifying partition-allocation ambiguity.} We next formalize the quantities used in the LAMA analysis reported in the
main paper (\cref{tab:lama_allocation_analysis}). Let
\begin{equation}
\boldsymbol{\pi}^\star(q)
=
\left[
\pi_1^\star(q),\ldots,\pi_N^\star(q),\pi_c^\star(q)
\right]
\end{equation}
denote the mask-derived target distribution. The corresponding
realized distributions before and after LAMA are denoted by
$\tilde{\boldsymbol{\pi}}^{\mathrm{raw}}(q)$ and
$\tilde{\boldsymbol{\pi}}^{\mathrm{LAMA}}(q)$, respectively.
Below, $\tilde{\boldsymbol{\pi}}(q)$ denotes either realized
distribution when defining metrics common to both stages.

For a query whose interior is unambiguously assigned to style $i$,
the realized probability mass can be decomposed as
\begin{equation}
\tilde{\pi}_i(q)
+
\tilde{\pi}_c(q)
+
\underbrace{
\sum_{k\neq i}
\tilde{\pi}_k(q)
}_{\mathcal{L}_{\mathrm{style}}(q)}
=
1
\label{eq:app_mass_decomposition}
\end{equation}
where
$\mathcal{L}_{\mathrm{style}}(q)$ denotes cross-style leakage.
It is the probability mass assigned to style partitions that are not
designated for the current query. Under the ideal mask-derived
allocation, this quantity is zero.

In the main paper (\cref{tab:lama_allocation_analysis}), interior statistics use
queries satisfying $M_i(q)\geq0.99$ and
$\sum_{k\neq i}M_k(q)\leq0.01$, whereas TV and JSD are evaluated over
all valid queries.

To measure allocation error over all valid queries, we first use total
variation distance:
\begin{equation}
D_{\mathrm{TV}}(q)
=
\frac{1}{2}
\sum_{g\in\{1,\ldots,N,c\}}
\left|
\tilde{\pi}_g(q)
-
\pi_g^\star(q)
\right|.
\label{eq:app_lama_tv}
\end{equation}
TV has a direct mass interpretation: it measures the total amount of
probability mass that must be reassigned to transform the realized
allocation into the target allocation.

We additionally use Jensen--Shannon divergence (JSD) to measure the
discrepancy between the complete realized and target partition
distributions. Defining
\begin{equation}
\boldsymbol{\pi}^{\mathrm{mix}}(q)
=
\frac{1}{2}
\left(
\tilde{\boldsymbol{\pi}}(q)
+
\boldsymbol{\pi}^\star(q)
\right)
\end{equation}
we compute
\begin{equation}
\begin{aligned}
D_{\mathrm{JS}}(q)
&=
\frac{1}{2}
D_{\mathrm{KL}}
\left(
\tilde{\boldsymbol{\pi}}(q)
\,\|\,\boldsymbol{\pi}^{\mathrm{mix}}(q)
\right)
\\
&\quad+
\frac{1}{2}
D_{\mathrm{KL}}
\left(
\boldsymbol{\pi}^\star(q)
\,\|\,\boldsymbol{\pi}^{\mathrm{mix}}(q)
\right)
\end{aligned}
\label{eq:app_lama_jsd}
\end{equation}

JSD is particularly suitable because zero-mass partitions arise
naturally from regional assignment. For example, in the two-style
setting, a query fully assigned to Style~1 with $\pi^\star=0.9$ has
the target distribution
$\boldsymbol{\pi}^\star(q)=[0.9,\,0,\,0.1]$
over Style~1, Style~2, and content, respectively. The zero mass on
Style~2 expresses that this style should not contribute at that query;
more generally, inactive style partitions receive zero target mass in
the $N$-style setting.

A direct KL divergence from the realized distribution to this target
would become infinite whenever nonzero attention leaks into a
zero-target partition. JSD instead compares both distributions with
$\boldsymbol{\pi}^{\mathrm{mix}}(q)$, and therefore remains finite and
symmetric even when their supports differ. This makes it suitable for
quantifying allocation mismatch in the presence of cross-style leakage.

Together, these quantities distinguish two aspects of the ambiguity:
the interior statistics reveal \emph{where the probability mass goes}
when a style assignment is unambiguous, whereas TV and JSD quantify
\emph{how far the full query-wise allocation is from the prescribed
distribution}. The near-zero TV and JSD after LAMA in the main paper
(\cref{tab:lama_allocation_analysis}) are therefore the empirical
counterpart of the exact realization derived in
\cref{eq:app_exact_allocation}.

\noindent {\bf Selection of $\pi^\star$.} The preceding derivation addresses the first question: given a valid
target distribution, LAMA realizes that allocation independently at
each query. A separate question is how the maximum style-mass budget
$\pi^\star$ should be chosen.

At a query fully assigned to a single style, the ideal target allocation
reduces to
\begin{equation}
\left[
\pi^\star,\,
1-\pi^\star
\right]
\end{equation}
over the assigned style and content partitions. Increasing
$\pi^\star$ therefore does not simply amplify style logits; it directly
transfers probability mass from content to style. Consequently,
$\pi^\star$ specifies an explicit operating point between regional
style fidelity and content preservation.

\begin{table}[t]
\centering

\small
\setlength{\tabcolsep}{5pt}
\renewcommand{\arraystretch}{1.08}
\begin{tabular}{c|cccc}
\hline
$\pi^\star$
& ArtFID $\downarrow$
& FID $\downarrow$
& LPIPS $\downarrow$
& R-FID $\downarrow$ \\
\hline
0.30
& 28.920
& 22.308
& \textbf{0.241}
& 30.472 \\
0.45
& 27.429
& 20.601
& \underline{0.270}
& 28.425 \\
0.60
& 24.827
& 17.816
& 0.319
& 24.984 \\
0.75
& 21.035
& 14.066
& 0.396
& 22.470 \\
0.90
& \textbf{14.780}
& \textbf{8.656}
& 0.531
& \textbf{11.260} \\
1.00
& \underline{15.235}
& \underline{8.890}
& 0.540
& \underline{11.344} \\
\hline
\end{tabular}
\caption{Quantitative evaluation in the two-style setting across
different values of $\pi^\star$.}
\label{tab:lama_pi_analysis}
\end{table}

As shown in \cref{tab:lama_pi_analysis}, increasing
$\pi^\star$ from $0.30$ to $0.90$ progressively improves FID and
R-FID, indicating stronger global and regional style fidelity, while
LPIPS increases as the content allocation decreases. The sweep
therefore exposes the expected style--content trade-off produced by
the target mass budget rather than merely reflecting a change in an
arbitrary logit scale.

The endpoint $\pi^\star=1.0$ is qualitatively different from the
preceding settings: in the ideal allocation, no probability mass
remains for content at a fully assigned query. Although this maximizes
the style budget, ArtFID, FID, LPIPS, and R-FID all degrade relative
to $\pi^\star=0.9$. Thus, completely removing the content reserve does
not improve either stylization quality or content preservation. We use
$\pi^\star=0.9$ in all experiments, which achieves the best ArtFID,
FID, and R-FID while retaining a nonzero content contribution.

\subsection{Sharpness reference and scaling property}
\label{appendix:tau}
\begin{figure}[t]
\centering
\includegraphics[width=0.75\linewidth]{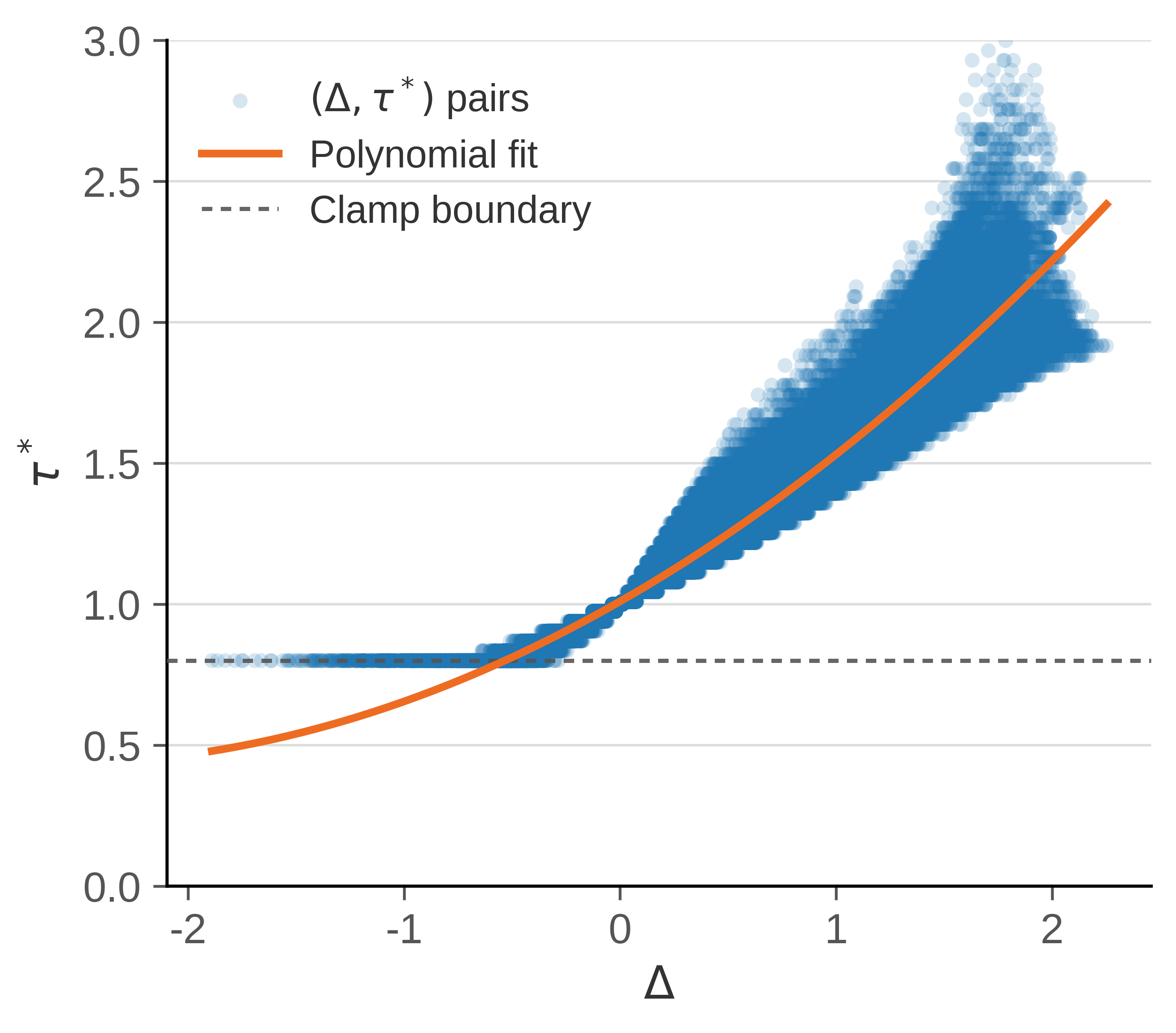}
\caption{
Empirical mapping from the sharpness gap $\Delta$ to the reference-matching
factor $\tau^\star$. Points denote head-level observations, the red curve denotes
the fitted quadratic function, and the dashed line marks the lower boundary
$\tau=0.8$ of the calibration grid.
}
\label{fig:delta_tau_mapping}
\end{figure}

STS uses content self-attention as an input-conditioned sharpness reference
for estimating the sharpening required for each attention head. Unlike a fixed
target, content self-attention reflects the native selectivity of the
corresponding content, layer, timestep, and head, while remaining unaffected
by the injected style references.

Throughout this appendix, $L\in\mathbb{R}^{N_q\times K}$ denotes the
query-by-key logit matrix of a single attention head, where $K$ is the number
of keys. In particular,
$L_c\in\mathbb{R}^{N_q\times N_k}$ and
$L_{\mathrm{LAMA}}\in
\mathbb{R}^{N_q\times (N+1)N_k}$.
Using the query-wise score in the main paper (\cref{eq:log_pmax}), the head-wise score
introduced in \cref{sec:sts} is

\begin{equation}
\log p_{\max}(L)
=
\frac{1}{N_q}
\sum_{q=1}^{N_q}
\log p_{\max}(L_{q,:}).
\label{eq:app_head_sharpness}
\end{equation}

The corresponding sharpness gap is defined in the main paper
(\cref{eq:sharpness_gap}). For a fixed logit vector
$\ell\in\mathbb{R}^{K}$ and $\tau>0$, let
$p_\tau=\operatorname{Softmax}(\tau\ell)$. Multiplicative scaling satisfies

\begin{equation}
\frac{d}{d\tau}\log p_{\max}(\tau\ell)
=
\ell_{\max}
-
\mathbb{E}_{p_\tau}[\ell]
\geq 0,
\label{eq:app_scaling_property}
\end{equation}

where $\ell_{\max}=\max_j\ell_j$ and
$\mathbb{E}_{p_\tau}[\ell]=\sum_j p_{\tau,j}\ell_j$.
Thus, increasing $\tau$ cannot decrease the peak probability of any query,
and averaging over queries preserves this monotonicity.

Let $\Delta(\tau)$ denote the residual sharpness gap after scaling. The ideal
reference-matching factor is

\begin{align}
\Delta(\tau)
&=
\log p_{\max}(L_c)
-
\log p_{\max}(\tau L_{\mathrm{LAMA}}),
\nonumber\\
\tau^\star
&=
\operatorname*{arg\,min}_{\tau>0}
\left|\Delta(\tau)\right|.
\label{eq:app_reference_factor}
\end{align}

Here, $\Delta(1)$ equals the unscaled gap $\Delta$ in
the main paper (\cref{eq:sharpness_gap}). Solving this optimization at every head,
layer, and timestep would require repeated evaluations. We therefore
predict $\tau$ from $\Delta$ using the empirical mapping described in the
following subsection.

\subsection{Empirical mapping construction}
\label{appendix:sts_mapping}

We construct an empirical mapping in the two-style setting from the unscaled sharpness gap $\Delta$ to
the reference-matching factor $\tau^\star$ using $200$ calibration samples drawn
from content and style images disjoint from those used for evaluation.
Collecting logits across attention heads, controlled layers, and diffusion
timesteps yields approximately $480{,}000$ head-level observations. For each
observation, we approximate \cref{eq:app_reference_factor} over $120$
uniformly spaced candidate values of $\tau$ in $[0.8,5]$ and record the pair
$(\Delta,\tau^\star)$.

As shown in \cref{fig:delta_tau_mapping}, $\Delta$ and $\tau^\star$ exhibit
an overall increasing relationship. However, observations with similar gaps
can require different factors because their complete logit spectra differ.
We therefore use the fitted function as an efficient approximation rather than
an exact analytical mapping.

We compare polynomial functions of different degrees using the coefficient of
determination. As shown in \cref{tab:poly_r2}, improvements become
marginal beyond the second order. We therefore use

\begin{equation}
f(\Delta)
=
a\Delta^2+b\Delta+c,
\label{eq:appendix_quadratic_mapping}
\end{equation}

where $a=0.08395$, $b=0.43705$, and $c=1.00998$. During inference, the
head-wise factor is computed as

\begin{equation}
\tau
=
\operatorname{clip}
\left(
f(\Delta),1,5
\right).
\label{eq:appendix_tau_clipping}
\end{equation}

The calibration grid includes factors below $1$ to obtain reference-matching
targets, whereas inference restricts $\tau$ to $[1,5]$ to prevent flattening
and excessive sharpening.

\begin{table}[t]
\centering

\small
\setlength{\tabcolsep}{5pt}
\renewcommand{\arraystretch}{1.08}
\begin{tabular}{cc}
\hline
Degree & $R^2$ $\uparrow$ \\
\hline
$1$ & $0.927$ \\
$2$ & $0.932$ \\
$3$ & $0.934$ \\
$4$ & $0.935$ \\
\hline
\end{tabular}
\caption{
Goodness of fit for polynomial mappings of different degrees.
}
\label{tab:poly_r2}
\end{table}

\noindent{\bf Head-wise estimation.}
We estimate one factor for each attention head because every head forms and
normalizes its own attention distribution. Heads within the same layer and
timestep can therefore exhibit different logit scales and sharpness deficits.
Sharing a single factor across these heads can lead to insufficient or
excessive sharpening.

Conversely, estimating a separate factor for every query can be sensitive to
local logit fluctuations. We therefore average the query-wise sharpness scores
within each head and apply the resulting $\tau$ to all queries in that head.
This estimation is repeated independently at every controlled layer and
diffusion timestep.

\subsection{Distributional effects of STS}
\label{appendix:sts_interaction}

We analyze the distributional effects of STS from two complementary
perspectives: the recovery of attention sharpness and the preservation of the
allocation prior established by LAMA.

\noindent{\bf Entropy-based verification of sharpness recovery.}
STS estimates $\tau$ using log-peak sharpness. To examine whether this
correction also improves the overall attention distribution, we additionally
measure entropy, which is not used to construct the $\Delta$-to-$\tau$
mapping. Consistent with the single-head notation above, we define

\begin{equation}
\begin{aligned}
P(L)
&=
\operatorname{Softmax}(L),\\
\mathcal{H}(L)
&=
-\frac{1}{N_q}
\sum_{q=1}^{N_q}
\sum_j
P(L)_{qj}\log P(L)_{qj}.
\end{aligned}
\label{eq:appendix_attention_entropy}
\end{equation}

Lower entropy indicates that probability is concentrated on fewer keys and
thus represents sharper attention. We compare $\mathcal{H}(L_c)$,
$\mathcal{H}(L_{\mathrm{LAMA}})$, and
$\mathcal{H}(\tau L_{\mathrm{LAMA}})$ in the four-style setting.

\begin{figure}[t]
\centering
\includegraphics[width=\linewidth]{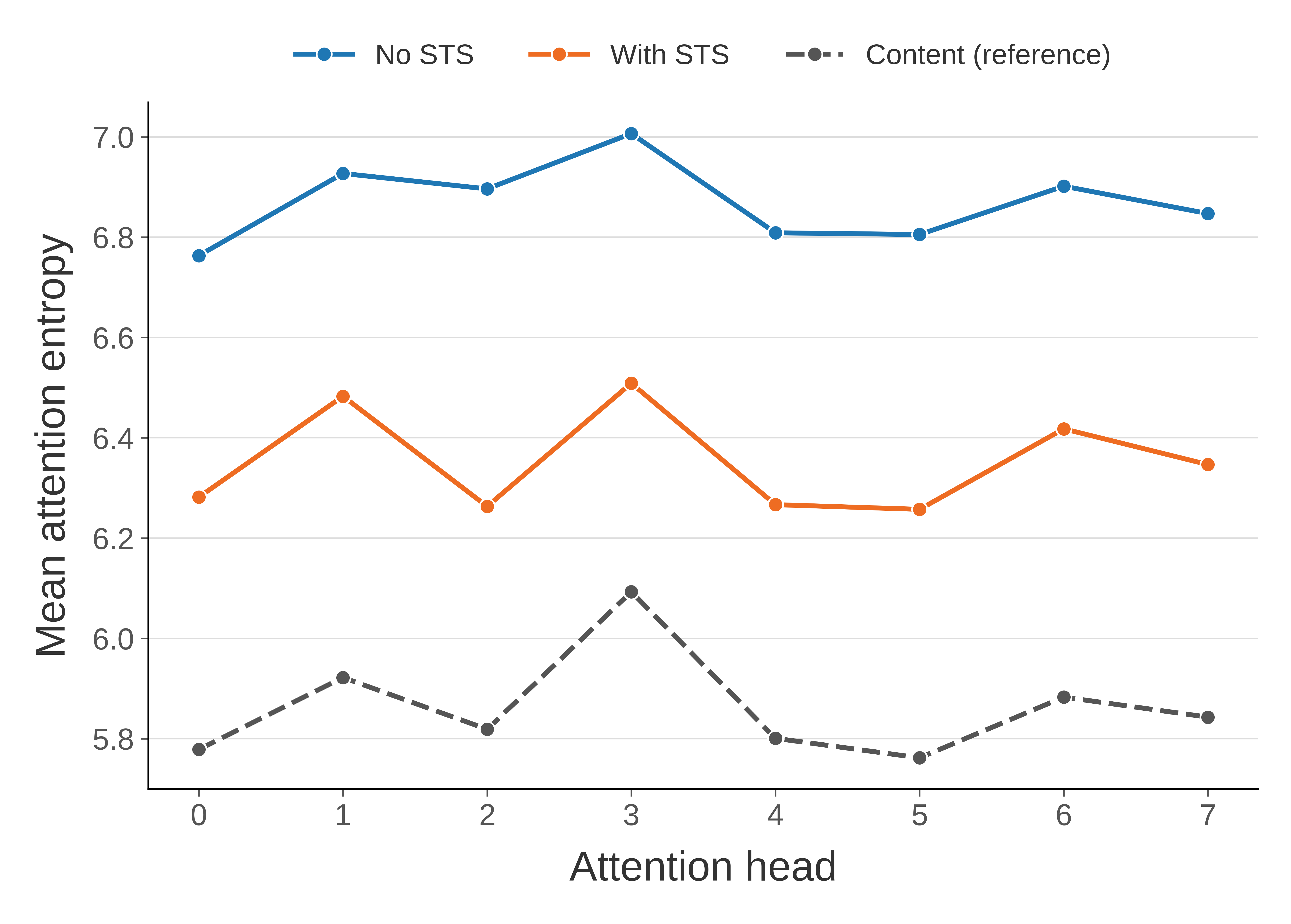}
\caption{
\textbf{Entropy-based verification of STS in the four-style setting.}
Mean attention entropy for content self-attention, the LAMA-adjusted joint
attention before STS, and the resulting attention after STS. Values for each
head are averaged over spatial queries, controlled layers, and $50$ DDIM
timesteps. Lower entropy indicates sharper attention.
}
\label{fig:sts_entropy_headwise}
\end{figure}

As shown in \cref{fig:sts_entropy_headwise}, the LAMA-adjusted joint
attention has consistently higher entropy than content self-attention, whereas
STS moves the joint distribution toward the content reference across all
heads. Because the before- and after-STS distributions share the same joint
key support, their entropy reduction directly reflects increased attention
concentration. Content self-attention is shown as a sharpness reference rather
than as an identical target distribution.

STS does not exactly match the reference entropy. This is expected because
$\tau$ is approximated from the log-peak sharpness gap, while entropy depends
on the complete probability distribution; matching the peak probability does
not necessarily reproduce the full distribution. Direct entropy matching
would require searching for $\tau$ through repeated Softmax evaluations at
every controlled head, layer, and timestep. STS instead provides an efficient
adaptive approximation without per-instance search, leaving more exact
distribution-level calibration as a possible extension.

Having verified the sharpness correction, we next examine how the subsequent
logit scaling affects the allocation prior established by LAMA.

\noindent{\bf Effect of STS on LAMA-guided allocation.}
STS does not exactly preserve the partition masses established by LAMA, since
subsequent logit scaling can modify their relative probabilities. We therefore
treat the LAMA targets as a mask-guided allocation prior and evaluate both the
change in partition mass and the retention of the query-wise allocation
pattern after STS.

Let $\xi$ index an evaluated sample, controlled layer, timestep, and attention
head. For partition $m\in\{c,1,\ldots,N\}$, its post-STS mass is

\begin{equation}
\tilde{\pi}_{m,\xi}^{\mathrm{STS}}(q)
=
\sum_{j\in\mathcal{S}_m}
A_{\mathrm{STS},\xi}(q,j),
\label{eq:appendix_post_sts_mass}
\end{equation}

where $\mathcal{S}_m$ denotes the key indices belonging to partition $m$. We
measure the mean total variation from the LAMA targets as

\begin{equation}
D_{\mathrm{mass}}
=
\frac{1}{2}
\mathbb{E}_{\xi,q}
\left[
\sum_{m\in\{c,1,\ldots,N\}}
\left|
\tilde{\pi}_{m,\xi}^{\mathrm{STS}}(q)
-
\pi_{m,\xi}^\star(q)
\right|
\right].
\label{eq:appendix_mass_deviation}
\end{equation}

We additionally measure the preservation of the query-wise allocation pattern
using

\begin{equation}
\rho_{\min}
=
\min_{i\in\{1,\ldots,N\}}
\mathbb{E}_{\xi}
\left[
\operatorname{Corr}_q
\left(
\pi_{i,\xi}^\star(q),
\widehat{\pi}_{i,\xi}^{\mathrm{STS}}(q)
\right)
\right],
\label{eq:appendix_mass_correlation}
\end{equation}

where $\operatorname{Corr}_q$ denotes the Pearson correlation computed across spatial
queries. Accordingly, $\rho_{\min}$ reports the lowest correlation among the
style partitions.

\begin{table}[t]
\centering

\small
\setlength{\tabcolsep}{5pt}
\renewcommand{\arraystretch}{1.08}

\begin{tabular}{cccc}
\hline
$N$
& $\mathbb{E}[\widehat{\pi}_c^{\mathrm{STS}}]$
& $D_{\mathrm{mass}}\downarrow$
& $\rho_{\min}\uparrow$ \\
\hline
$2$ & $0.080$ & $0.038$ & $0.995$ \\
$3$ & $0.078$ & $0.035$ & $0.998$ \\
$4$ & $0.078$ & $0.035$ & $0.995$ \\
$5$ & $0.078$ & $0.036$ & $0.993$ \\
\hline
\end{tabular}
\caption{
Effect of STS on the LAMA-guided allocation.
The mean LAMA target for the content partition is
$\mathbb{E}[\pi_c^\star]=0.100$.
$\rho_{\min}$ measures agreement with the mask-induced query-wise allocation.
}
\label{tab:sts_mass}
\end{table}

As shown in \cref{tab:sts_mass}, STS introduces a limited but nonzero
deviation from the LAMA targets, and the deviation remains consistent across
different numbers of styles. Meanwhile, the high query-wise correlation
indicates that the mask-induced allocation pattern is largely retained. In
other words, STS changes the exact partition masses but largely preserves
where each style is favored. LAMA therefore provides the regional allocation
prior upon which STS refines the selectivity of the joint attention
distribution.

\subsection{High-pass filtering in DDI and sensitivity analysis}
\label{appendix:high}

This section specifies the Gaussian high-pass operator
$\mathcal{H}_r$ used to obtain $F_c^{\mathrm{high}}$ in DDI and examines
the effect of the normalized Gaussian frequency scale $r$.

\noindent {\bf Gaussian high-pass mask.}
Let $F_c\in\mathbb{R}^{B\times C\times H\times W}$ denote the
residual-branch output obtained from the corresponding content-path
ResBlock. DDI is applied to the last ResBlock of decoder output blocks
7--12 at all denoising timesteps. Blocks 7--9 operate at a spatial
resolution of $32\times32$, whereas blocks 10--12 operate at
$64\times64$.

For a frequency coordinate $(u,v)$ in the centered frequency plane,
its radial distance from the center is defined as
\begin{equation}
D(u,v)
=
\sqrt{
\left(u-\left\lfloor H/2\right\rfloor\right)^2
+
\left(v-\left\lfloor W/2\right\rfloor\right)^2
}.
\label{eq:frequency_distance}
\end{equation}
The normalized Gaussian frequency scale $r$ is converted into
frequency-bin units as
\begin{equation}
\sigma_r=r\min(H,W).
\label{eq:gaussian_scale}
\end{equation}
The Gaussian high-pass mask is then defined as
\begin{equation}
M_{\text{gauss-high}}(r;u,v)
=
1-\exp\left(
-\frac{D(u,v)^2}
{2\left(\sigma_r^2+\epsilon\right)}
\right),
\label{eq:gaussian_high_pass}
\end{equation}
where $\epsilon=10^{-8}$ ensures numerical stability. The mask
vanishes at the frequency center and increases smoothly with radial
distance. Increasing $r$ enlarges $\sigma_r$, suppressing a broader
frequency range and leaving a narrower effective high-frequency
passband. Thus, $r$ controls a smooth Gaussian frequency scale rather
than a hard cutoff.

\noindent {\bf High-frequency feature extraction.}
Let $\mathcal{F}$ denote the orthonormally normalized 2D Fourier
transform followed by a center shift, and let $\mathcal{F}^{-1}$ denote
the corresponding inverse operation, consisting of an inverse shift
followed by the inverse Fourier transform. Both are applied along the
spatial dimensions. The Gaussian high-pass operator is defined as
\begin{equation}
F_c^{\mathrm{high}}
=
\mathcal{H}_r(F_c)
:=
\mathcal{F}^{-1}
\left(
\mathcal{F}(F_c)
\odot
M_{\text{gauss-high}}(r)
\right),
\label{eq:appendix_high_frequency}
\end{equation}
where $\odot$ denotes element-wise multiplication, and the same mask is
broadcast across the batch and channel dimensions. Since $F_c$ is
real-valued and the Gaussian mask preserves conjugate symmetry, the
inverse-transform output is real-valued in exact arithmetic; in
implementation, we retain its real component to remove minor numerical
imaginary residuals. The resulting $F_c^{\mathrm{high}}$ emphasizes
content responses associated with edges, textures, and other fine
structures.

As specified in the main paper (\cref{eq:ddi}), $\omega$ is computed separately for
each sample from the cosine discrepancy between $\Delta F_x$ and $F_c$
after flattening their channel and spatial dimensions. It is reshaped
to $B\times1\times1\times1$ and broadcast when modulating
$F_c^{\mathrm{high}}$.

\begin{figure}[t]
\centering
\includegraphics[width=0.8\linewidth]{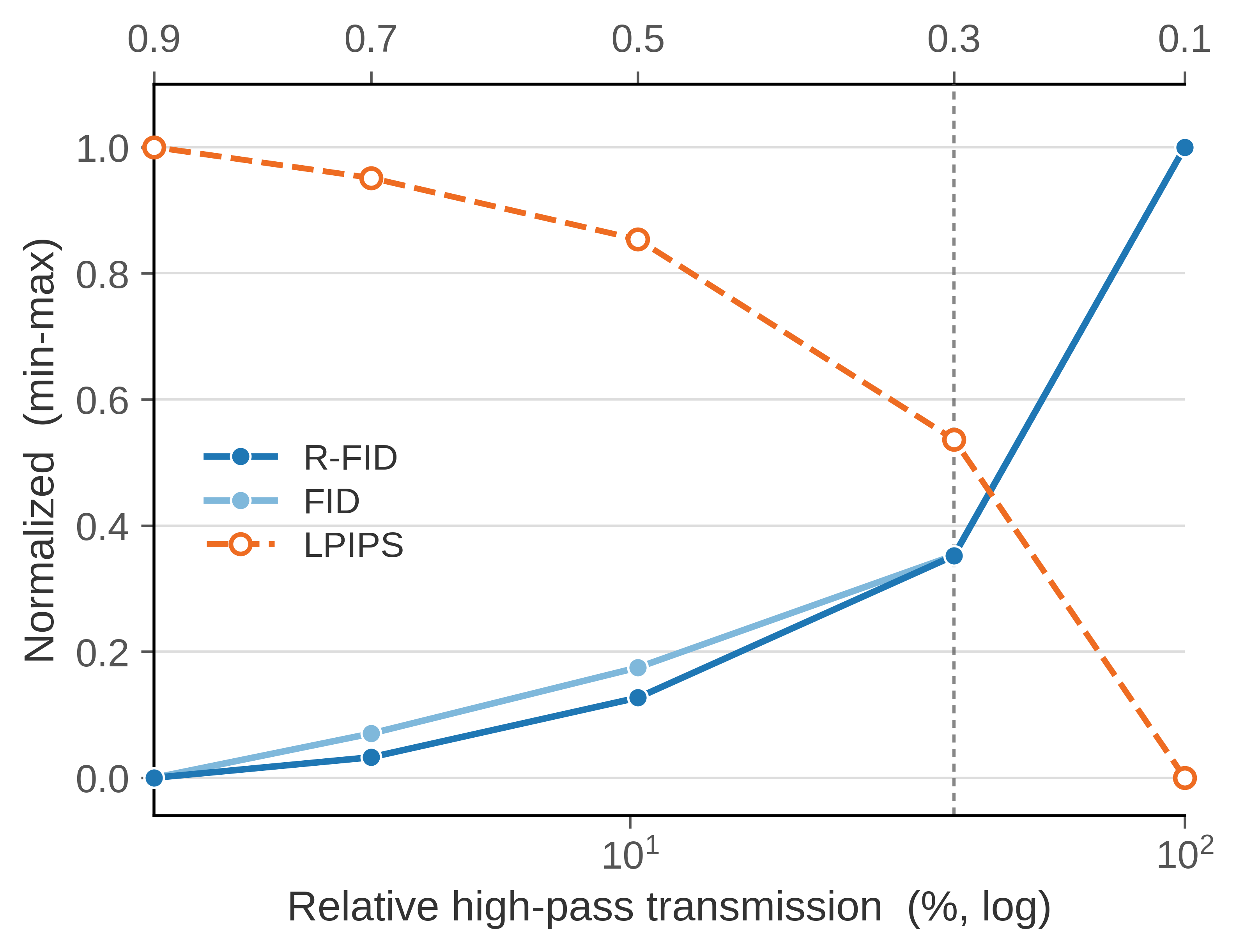}
\caption{Sensitivity of DDI to the Gaussian frequency scale $r$.
The independently min--max-normalized metrics ($0$ denotes best) are
plotted against the relative squared-gain transmission $T(r)$ on the
lower logarithmic axis; the upper axis reports $r$, and the dashed line
marks the default $r=0.3$.
}
\label{fig:ratio_sweep}
\end{figure}
\begin{table}[t]
\centering

\small
\setlength{\tabcolsep}{5pt}
\renewcommand{\arraystretch}{1.08}
\begin{tabular}{lcccc}
\hline
$r$ & ArtFID $\downarrow$
& FID $\downarrow$
& LPIPS $\downarrow$
& R-FID $\downarrow$ \\
\hline
0.1 & 15.276 & 9.126 & \textbf{0.509} & 12.055 \\
0.3 & 14.780 & 8.656 & \underline{0.531} & 11.260 \\
0.5 & 14.707 & 8.527 & 0.544 & 10.984 \\
0.7 & \underline{14.626} & \underline{8.451} & 0.548 & \underline{10.868} \\
0.9 & \textbf{14.570} & \textbf{8.400} & 0.550 & \textbf{10.828} \\
\hline
\end{tabular}
\caption{Sensitivity analysis of the normalized Gaussian frequency scale $r$ in the two-style setting. Lower values indicate better performance.}
\label{tab:r_sensitivity}
\end{table}

\noindent {\bf Sensitivity to frequency scale $r$.}
We evaluate the normalized Gaussian frequency scale
$r\in\{0.1,0.3,0.5,0.7,0.9\}$ in \cref{fig:ratio_sweep} and \cref{tab:r_sensitivity}.
To relate $r$ to the nominal filtering strength, the figure plots
independently min--max-normalized FID, R-FID, and LPIPS against the
relative squared-gain transmission
$T(r)=\overline{M_r^2}/\overline{M_{0.1}^2}$, where
$M_r=M_{\text{gauss-high}}(r)$ and the overline denotes the mean over
frequency bins. This is a filter-only quantity rather than the actual
injected feature energy, which additionally depends on the spectrum of
$F_c$ and the sample-dependent weight $\omega$.
The transmission values at $32\times32$ and $64\times64$ differ by less
than $0.05$ percentage points, making a shared rounded transmission
axis sufficient. As $r$ increases, the relative transmission decreases
from $100\%$ at $r=0.1$ to approximately $38\%$ at $r=0.3$ and
$1.4\%$ at $r=0.9$.

Although $r=0.1$ achieves the best LPIPS and $r=0.9$ yields the best
ArtFID, FID, and R-FID, the latter provides only $1.4\%$ relative
squared-gain transmission, leaving little nominal high-pass
transmission for DDI's intended detail compensation. In contrast,
$r=0.3$ retains approximately $38\%$ transmission while capturing
$65\%$ of the total R-FID reduction from $r=0.1$ to $r=0.9$ and
incurring $54\%$ of the corresponding LPIPS increase. It therefore
provides a practical operating point near the knee of the trade-off,
and we adopt $r=0.3$ as the default.

\subsection{Region-FID}
\label{app:region_fid}

\noindent {\bf Motivation.}
As stated in \cref{sec:lama}, MAST receives $N$ style--mask pairs
$\{(I_s^{(i)},M_i)\}_{i=1}^{N}$, where $M_i$ specifies the target
region of style $i$.
FID measures distributional similarity to the style images, but it cannot
verify whether each region matches its assigned style.
The most direct regional extension is to pool the stylized output only within
$M_i$ while retaining FID's ordinary global pooling for the assigned style
image $I_s^{(i)}$; we call this naive extension masked FID (M-FID).

Let $\mathcal{P}$ be the spatial index set of the FID encoder's pre-pooling
feature map and $\mathcal{P}_i\subset\mathcal{P}$ the cells covered by $M_i$.
With $F_p^o$ and $F_p^{s,(i)}$ denoting output and style features at cell $p$,
M-FID forms
\begin{equation*}
\begin{gathered}
z_i^o
= \frac{1}{|\mathcal{P}_i|}
  \sum_{p\in\mathcal{P}_i} F_p^o,
\qquad
z_i^{s,\mathrm{full}}
= \frac{1}{|\mathcal{P}|}
  \sum_{p\in\mathcal{P}} F_p^{s,(i)},\\
\mathrm{M\!-\!FID}
= \mathrm{FID}\!\left(
\mathbf{\Omega}^o,
\mathbf{\Omega}^{s,\mathrm{full}}
\right).
\end{gathered}
\end{equation*}

where
$\mathbf{\Omega}^o=\{z_i^o\}$ and
$\mathbf{\Omega}^{s,\mathrm{full}}
=\{z_i^{s,\mathrm{full}}\}$
collect all evaluated output--style assignments.
In other words, M-FID compares an average over regional support
$|\mathcal{P}_i|$ with an average over full-image support $|\mathcal{P}|$.

To isolate the consequence, suppose that the spatial features on both sides
are drawn from the same distribution with covariance $\Sigma$ and are
uncorrelated across cells.
Spatial averaging then yields
\begin{equation}
\begin{gathered}
\operatorname{Cov}(z_i^o)
  = \frac{\Sigma}{|\mathcal{P}_i|},\quad
\operatorname{Cov}(z_i^{s,\mathrm{full}})
  = \frac{\Sigma}{|\mathcal{P}|}, \\
\Delta_i^{\mathrm{cov}}
  = \left(
      \frac{1}{|\mathcal{P}_i|}
      -\frac{1}{|\mathcal{P}|}
     \right)\Sigma.
\label{eq:mfid_support_bias}
\end{gathered}
\end{equation}
Thus, $\Delta_i^{\mathrm{cov}}$ enters the covariance component of FID even when the underlying regional and style feature distributions are identical, and its magnitude grows as the output region becomes smaller, which occurs on average as a fixed-size output is partitioned among more styles.
For readability, \cref{eq:mfid_support_bias} uses binary support notation;
the area-preserving weights used in evaluation follow the same argument through
their effective pooled support.

\noindent {\bf Support-matched R-FID.}
R-FID removes this area-dependent mismatch by comparing embeddings with equal
spatial support.
For each output-region embedding, it constructs a reference embedding by
sampling the same effective support from the full feature map of the assigned
style image.
Sampling positions vary across Monte Carlo repeats, so only the amount of
supporting evidence is matched; the output mask's location and shape are never
imposed on the style image.
Denoting the output embedding collection by $\mathbf{\Omega}^o$ and the
reference collection from the $u$-th Monte Carlo repeat by
$\mathbf{\Omega}^{s,(u)}$, we compute
\begin{equation}
\mathrm{R\!-\!FID}
=
\frac{1}{R}
\sum_{u=1}^{R}
\mathrm{FID}\!\left(
\mathbf{\Omega}^o,
\mathbf{\Omega}^{s,(u)}
\right),
\qquad R=50.
\label{eq:rfid}
\end{equation}
Area-preserving mask weights retain fractional regional support, and the same
weights are used to pool the sampled style features.
\begin{table}[t]
\centering

\small
\setlength{\tabcolsep}{5pt}
\renewcommand{\arraystretch}{1.08}
\begin{tabular}{lcccc}
\hline
Metric & 2-style & 3-style & 4-style & 5-style \\
\hline
ArtFID &  0.881 &  0.929 &  0.952 &  0.952 \\
FID    &  0.881 &  0.881 &  0.881 &  0.881 \\
LPIPS  & $-0.690$ & $-0.762$ & $-0.738$ & $-0.571$ \\
\hline
\end{tabular}
\caption{
Spearman rank correlation between R-FID and complementary metrics across the
eight evaluated methods.
}
\label{tab:region_fid_correlation}
\end{table}

\begin{table}[t]
\centering

\small
\setlength{\tabcolsep}{5pt}
\renewcommand{\arraystretch}{1.08}
\begin{tabular}{lcccc}
\hline
Metric & 2 & 3 & 4 & 5 \\
\hline
M-FID  & 17.073 & 27.735 & 35.915 & 41.007 \\
ArtFID & 14.780 & 14.802 & 15.014 & 15.117 \\
FID    &  8.656 &  8.700 &  8.905 &  8.929 \\
R-FID  & 11.260 & 14.219 & 18.797 & 20.883 \\
\hline
\end{tabular}
\caption{
Comparison of M-FID, ArtFID, FID, and R-FID on MAST outputs across different
numbers of styles.
}
\label{tab:region_fid_support}
\end{table}

\begin{table}[t]
\centering

\small
\setlength{\tabcolsep}{5pt}
\renewcommand{\arraystretch}{1.08}
\begin{tabular}{c|c}
\hline
$\pi^\star$ & R-FID $\downarrow$ \\
\hline
0.10 & 71.291 \\
0.50 & 60.264 \\
0.90 & 18.797 \\
\hline
\end{tabular}
\caption{
Controlled style-weakening test in the four-style MAST setting.
Only $\pi^\star$ is varied.
}
\label{tab:region_fid_monotonicity}
\end{table}

\noindent {\bf Validation.}
We assess R-FID through three complementary properties.
First, we evaluate its consistency with
established metrics in \cref{tab:region_fid_correlation}.
Across the eight methods, R-FID has a Spearman correlation of
$0.881$--$0.952$ with ArtFID and $0.881$ with FID across the four style-count
settings.
These strong positive correlations show that, while R-FID evaluates assigned regions, its distributional style-fidelity ordering remains consistent with that of widely used metrics.
Its negative correlation with LPIPS reflects the expected content--style
trade-off rather than disagreement in style fidelity.

Second, we examine sensitivity to the decreasing average output-region size as the number of styles increases from two to five in \cref{tab:region_fid_support}. ArtFID and FID remain nearly stable, with ranges of $0.337$ and $0.273$, respectively, indicating that the global evaluation changes little across the four settings. In contrast, the direct region-to-global M-FID increases from $17.073$ to $41.007$, a range of $23.934$, whereas R-FID increases from $11.260$ to $20.883$, a substantially smaller range of $9.623$. The smaller R-FID range shows that support matching reduces the region-size-dependent inflation caused by the covariance mismatch derived in the Motivation, while retaining sensitivity to regional style fidelity.

Third, we test whether R-FID retains the
expected sensitivity of FID to style strength in \cref{tab:region_fid_monotonicity}.
When only the style-mass budget $\pi^\star$ is reduced from $0.90$ to $0.50$
and $0.10$, thereby weakening regional style injection, R-FID worsens
monotonically from $18.797$ to $60.264$ and $71.291$, respectively.
This controlled response shows that R-FID correctly detects increasing
discrepancy from the assigned style.

\subsection{Post-hoc compositing for \textbf{G-S} and \textbf{R-S} baselines}
\label{appendix:blending}

\noindent {\bf Compositing protocol.}
For each \textbf{G-S} or \textbf{R-S} baseline, let $\{(I_o^{(i)}, M_i)\}_{i=1}^{N}$ denote the ordered output--mask pairs generated using the corresponding style references $I_s^{(i)}$.
The binary masks form a mutually exclusive and exhaustive partition,
satisfying $\sum_{i=1}^{N} M_i(p)=1$ at every pixel $p$.
In addition to hard-mask compositing, we evaluate three widely used
post-hoc alternatives: alpha compositing~\cite{alpha}, Poisson
blending~\cite{poisson}, and Laplacian-pyramid
blending~\cite{laplacian}. The same fixed index order is used for Poisson and
Laplacian-pyramid blending.
\textbf{R-M} baselines and {\MAST} instead use their native multi-style inference
without post-hoc compositing.

\noindent {\bf Hard-mask compositing.}
Hard-mask compositing combines the per-style outputs according to their
assigned masks:
\begin{equation}
I_{\mathrm{hard}}(p)
=
\sum_{i=1}^{N}
M_i(p)I_o^{(i)}(p).
\label{eq:hard_compositing}
\end{equation}
Since the masks form a binary partition, this operation selects exactly
one output at each pixel without mixing, which can produce abrupt
transitions at region boundaries.

\noindent {\bf Alpha compositing.}
Alpha compositing~\cite{alpha} softens these boundary transitions by
feathering the masks before combining the per-style outputs. To avoid oversmoothing small regions, we adapt the radius to the binary mask area $a_i=\sum_p M_i(p)$ as
$\eta_i=\max\!\left(1,\left\lfloor
\min\!\left(8,\frac{1}{3}\sqrt{a_i/\pi}\right)
\right\rfloor\right)$.
Let $\mathcal{G}_{\eta_i}$ denote Gaussian blurring with kernel size
$(2\eta_i+1)\times(2\eta_i+1)$. The normalized alpha weights and
composite are
\begin{equation}
\begin{gathered}
\alpha_i(p)
=
\frac{\mathcal{G}_{\eta_i}(M_i)(p)}
{\sum_{j=1}^{N}\mathcal{G}_{\eta_j}(M_j)(p)},
\\
I_{\mathrm{alpha}}(p)
=
\sum_{i=1}^{N}
\alpha_i(p)I_o^{(i)}(p).
\end{gathered}
\label{eq:alpha_compositing}
\end{equation}
This softly mixes adjacent outputs near region boundaries.

\begin{table}
\centering

\footnotesize
\setlength{\tabcolsep}{4pt}
\renewcommand{\arraystretch}{1.08}
\begin{tabular}{llcccc}
\hline
Method & Strategy & $N=2$ & $N=3$ & $N=4$ & $N=5$ \\
\hline
$Z^\ast$
& Hard-mask & 17.311 & \textbf{17.009} & \textbf{17.253} & \textbf{16.926} \\
& Alpha & \textbf{17.215} & 17.381 & 17.602 & 17.530 \\
& Poisson & 17.476 & 17.484 & 17.324 & 17.295 \\
& Laplacian-pyramid & 17.265 & 17.449 & 17.436 & 17.680 \\
\hline
StyleID
& Hard-mask & 16.866 & 17.141 & 17.072 & 17.009 \\
& Alpha & 16.935 & 17.222 & 17.371 & 17.224 \\
& Poisson & 16.949 & 16.938 & 17.065 & \textbf{16.927} \\
& Laplacian-pyramid & \textbf{16.635} & \textbf{16.819} & \textbf{16.995} & 17.175 \\
\hline
CoCoDiff
& Hard-mask & 16.883 & 17.349 & \textbf{17.345} & \textbf{17.308} \\
& Alpha & 17.175 & 17.344 & 17.671 & 17.490 \\
& Poisson & 17.457 & 17.347 & 17.537 & 17.675 \\
& Laplacian-pyramid & \textbf{16.870} & \textbf{17.217} & 17.425 & 17.622 \\
\hline
Any-to-Any
& Hard-mask & 16.826 & 17.039 & 17.361 & 17.309 \\
& Alpha & 16.009 & 16.601 & 16.938 & 17.046 \\
& Poisson & 15.893 & 16.175 & \textbf{16.311} & \textbf{16.238} \\
& Laplacian-pyramid & \textbf{15.664} & \textbf{16.086} & 16.381 & 16.330 \\
\hline
BPC
& Hard-mask & \textbf{25.631} & \textbf{26.015} & \textbf{26.311} & 26.216 \\
& Alpha & 25.639 & 26.246 & 26.360 & \textbf{25.384} \\
& Poisson & 25.683 & 26.710 & 27.058 & 25.667 \\
& Laplacian-pyramid & 25.804 & 26.530 & 26.623 & 26.290 \\
\hline
\end{tabular}
\caption{ArtFID comparison of four post-hoc compositing strategies for
G-S and R-S baselines. 
Lower values are better, and bold indicates the strategy selected for each method and number of styles.
}
\label{tab:compositing_artfid}
\end{table}

\noindent {\bf Poisson blending.}
Poisson blending~\cite{poisson} reduces boundary discontinuities through
gradient-domain reconstruction. Conceptually, let $C_1=I_o^{(1)}$ and
$\Omega_i=\{p\mid M_i(p)=1\}$ denote the region inserted at step $i$.
For $i=2,\ldots,N$, we formulate
\begin{equation}
\begin{gathered}
C_i
=
\arg\min_X
\int_{\Omega_i}
\left\|
\nabla X(p)-\nabla I_o^{(i)}(p)
\right\|_2^2\,dp,
\\
X|_{\partial\Omega_i}
=
C_{i-1}|_{\partial\Omega_i}.
\end{gathered}
\label{eq:poisson_compositing}
\end{equation}
The fixed index order yields $I_{\mathrm{Poisson}}=C_N$.
In implementation, we use OpenCV's \texttt{NORMAL\_CLONE}, with the
center of the mask bounding box as the cloning center. Because each
region inherits its boundary condition from the current canvas, the
result is order-dependent: local source gradients are preserved, while
low-frequency color and tone can be influenced by the preceding
composite.

\noindent {\bf Laplacian-pyramid blending.}
Laplacian-pyramid blending~\cite{laplacian} reduces boundary
discontinuities by blending the per-style outputs across multiple
spatial scales. We initialize $C_1=I_o^{(1)}$ and sequentially composite
$I_o^{(i)}$ onto $C_{i-1}$ for $i=2,\ldots,N$. Each mask $M_i$ is smoothed with a Gaussian filter of $\sigma=2$
before constructing its Gaussian pyramid. Let $W_i^{(\ell)}$ denote
level $\ell$ of the resulting Gaussian mask pyramid, and let
$\mathcal{L}^{(\ell)}(I)$ denote the corresponding Laplacian image
level. We compute
\begin{equation}
\begin{gathered}
B_i^{(\ell)}
=
W_i^{(\ell)}
\odot
\mathcal{L}^{(\ell)}\!\left(I_o^{(i)}\right)
\\
+
\left(1-W_i^{(\ell)}\right)
\odot
\mathcal{L}^{(\ell)}(C_{i-1}).
\end{gathered}
\label{eq:laplacian_level_blending}
\end{equation}
We reconstruct $C_i$ from
$\{B_i^{(\ell)}\}_{\ell=0}^{L}$, using $L=7$ for the
$512\times512$ outputs. Applying this procedure in the fixed index order
yields $I_{\mathrm{Laplacian}}=C_N$.

\begin{figure}[t]
    \centering
    \includegraphics[width=\linewidth]{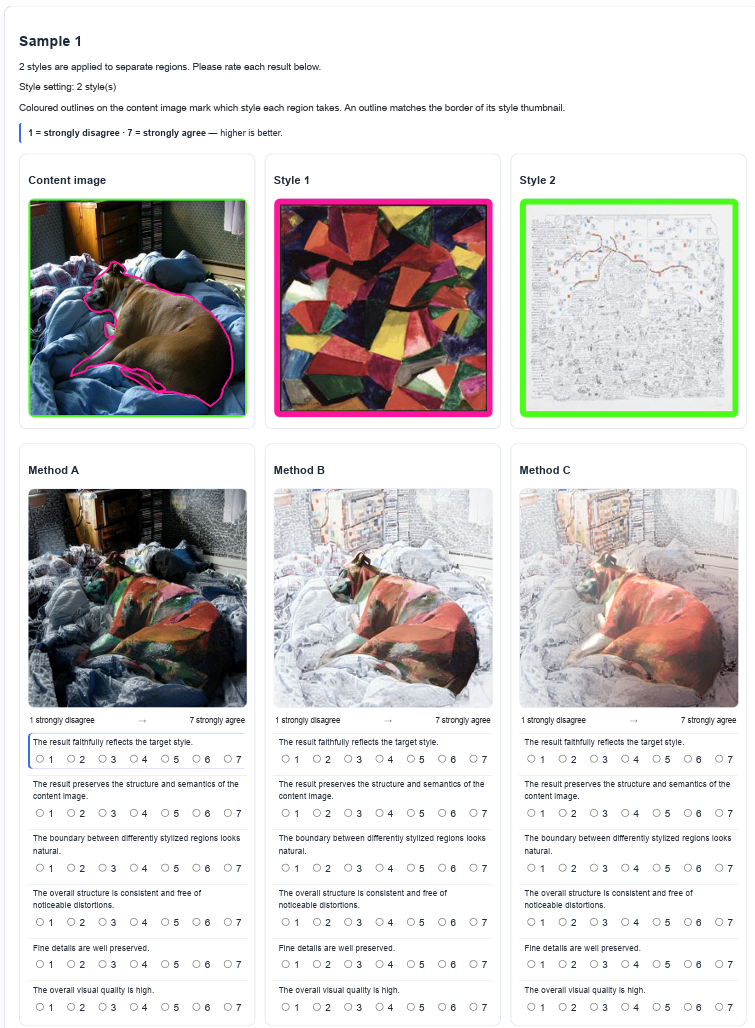}
    \caption{User-study interface showing the content image, style references, and method outputs rated using six 7-point Likert
    criteria.}
    \label{fig:user_study_interface}
\end{figure}

\noindent {\bf Dataset-level strategy selection.}
Let
$\mathcal{B}=
\{\mathrm{hard},\mathrm{alpha},\mathrm{Poisson},
\mathrm{Laplacian}\}$
denote the candidate strategies. For each \textbf{G-S} or \textbf{R-S} baseline $m$,
style count $N\in\{2,3,4,5\}$, and strategy $b\in\mathcal{B}$, let
$\mathcal{O}_{m,N}^{(b)}$ denote the complete evaluation output set.
We select a single strategy by dataset-level ArtFID:
\begin{equation}
b_{m,N}^{\star}
=
\arg\min_{b\in\mathcal{B}}
\operatorname{ArtFID}
\left(\mathcal{O}_{m,N}^{(b)}\right).
\label{eq:compositing_selection}
\end{equation}
The selected strategy is fixed across all samples within each $(m,N)$
setting, and all reported metrics are computed from the same output set
$\mathcal{O}_{m,N}^{(b_{m,N}^{\star})}$. This protocol gives each baseline its most favorable dataset-level
configuration under the primary metric.

\noindent {\bf Quantitative comparison.}
As shown in \cref{tab:compositing_artfid}, no single strategy is optimal across all settings. Among the 20 $(m,N)$ configurations, the hard-mask, Laplacian-pyramid, Poisson, and alpha strategies are
selected in 8, 7, 3, and 2 cases, respectively. Relative to using hard-mask compositing throughout, dataset-level
selection chooses an alternative strategy in 12 cases and reduces ArtFID by $0.301$ on average, with a maximum reduction of $1.162$ for Any-to-Any at $N=2$. The optimal strategy can also vary with $N$: Any-to-Any uses
Laplacian-pyramid blending at $N=2,3$ and Poisson blending at
$N=4,5$, supporting separate selection for each $(m,N)$ setting.

\begin{figure}[t]
\centering
    \includegraphics[width=\linewidth]{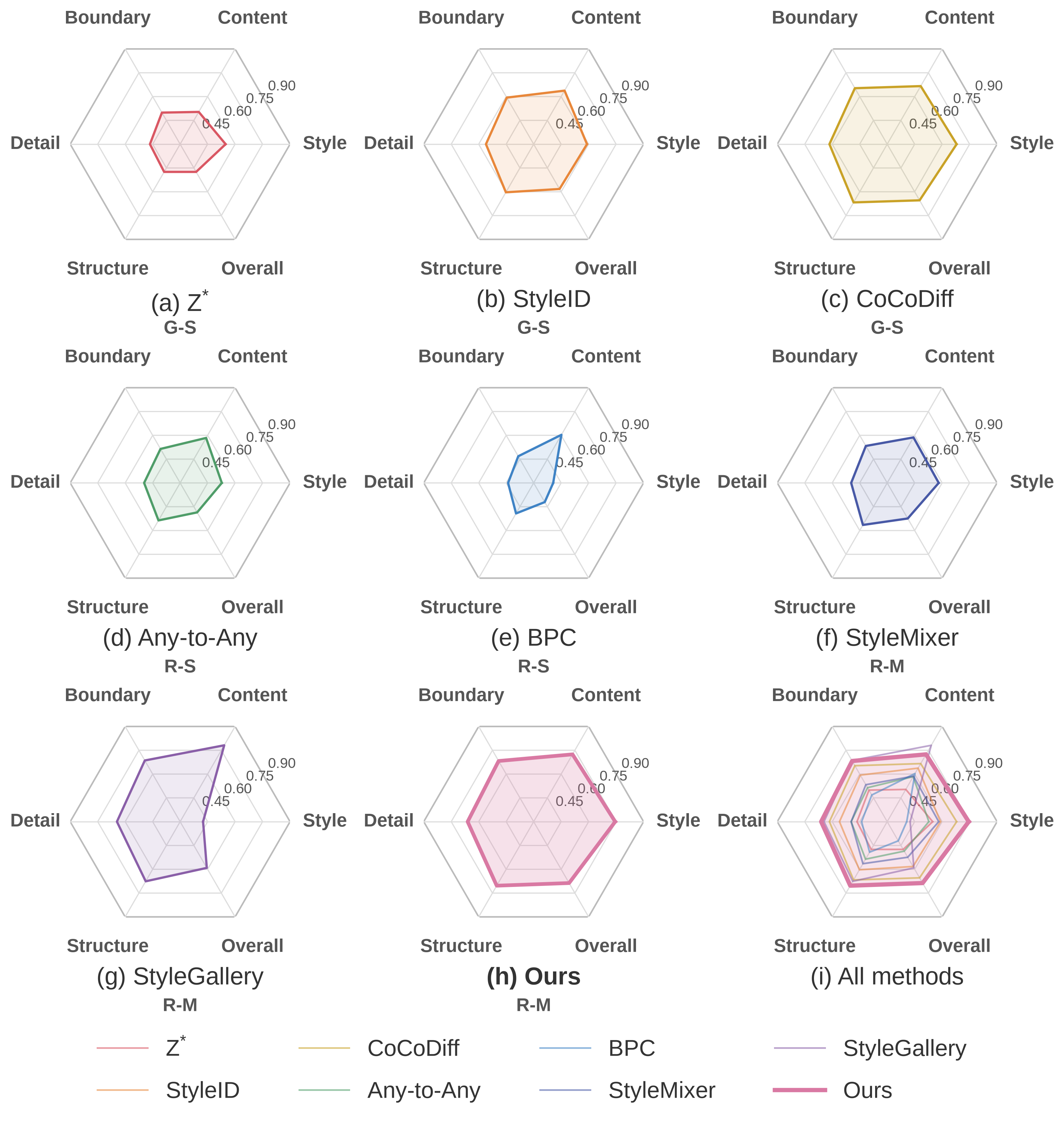}
    \caption{User study radar charts for eight methods. (a)--(h) show per-method average scores, and (i) overlays all methods. Mean Likert scores are divided by $7$; radial axes span $0.30$--$0.95$ to make differences visible.}
    \label{fig:us_result}
\end{figure}

\subsection{Additional stylization and evaluation results}

\begin{figure*}[t]

    \includegraphics[width=\linewidth]{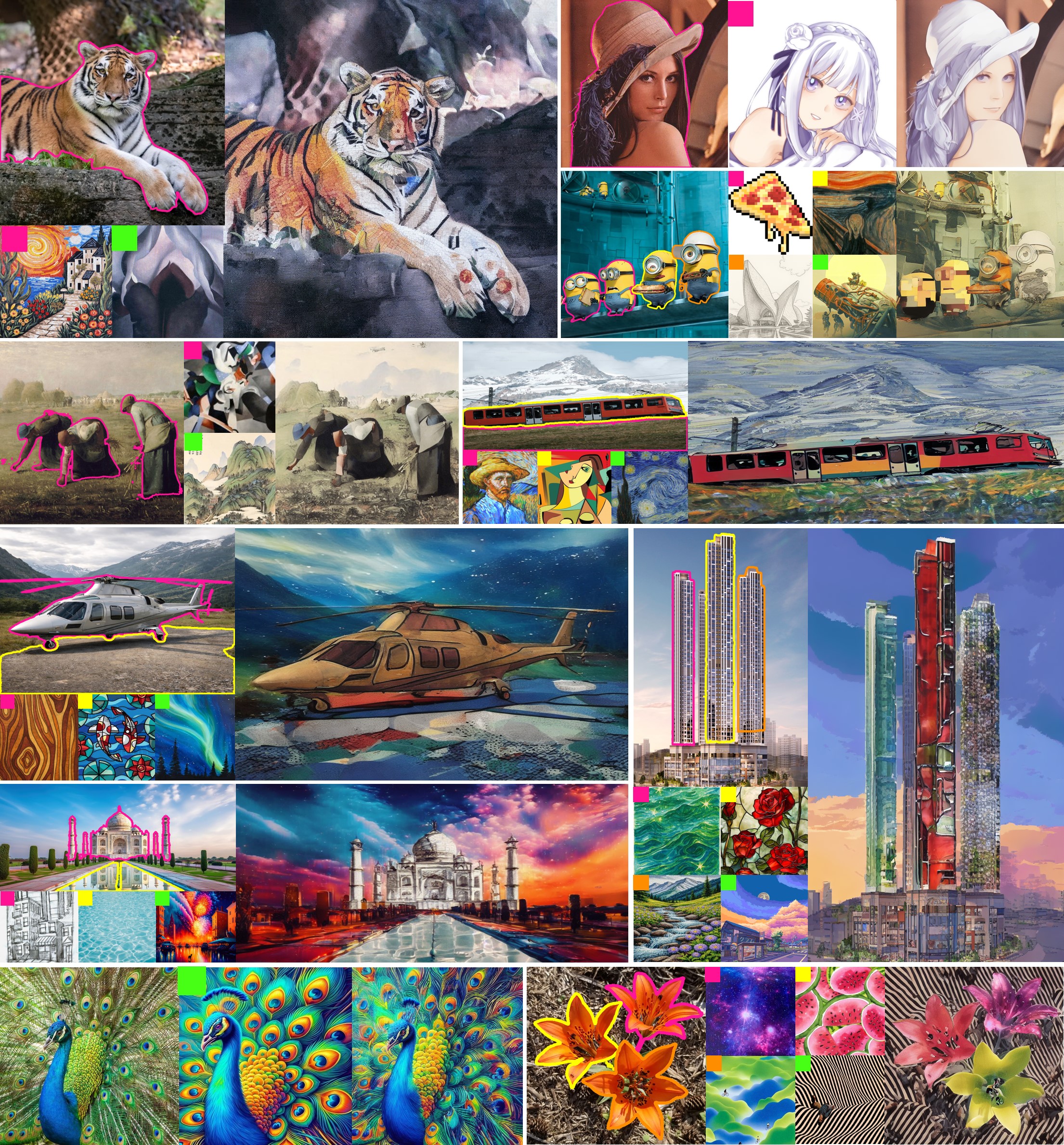}
    \caption{Qualitative scalability of {\MAST} across $1$--$4$ styles, multiple aspect ratios, and $512$--$1024$-pixel resolutions
(\textcolor{green}{green}: background).
    }

    \label{fig:app_ours_fig}
\end{figure*}

\begin{figure*}[t]
    \includegraphics[width=\linewidth]{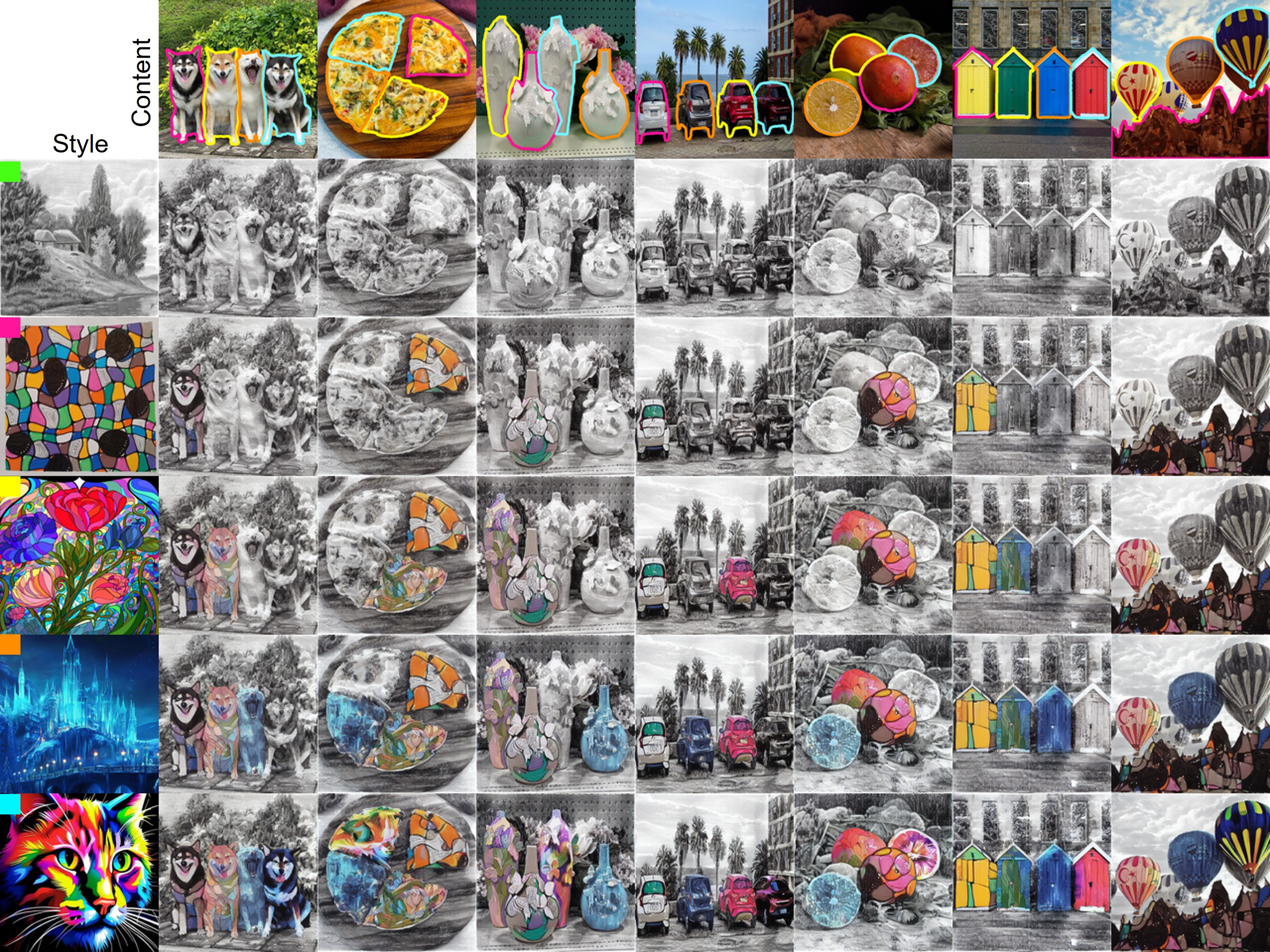}
    \caption{Progressive addition of up to five styles under a fixed five-region spatial partition
    (\textcolor{green}{green}: background).
    }
    \label{fig:ab_5sty}
\end{figure*}

\noindent {\bf User study.}
\label{app:user_study}
\cref{fig:user_study_interface} shows the interface.
We constructed four cases, one for each $N\in\{2,3,4,5\}$, and included
outputs from all eight methods in each case, yielding 32 images. Thirty
participants rated every output on a 7-point Likert scale using six criteria:
style faithfulness, content preservation, boundary naturalness, structural
consistency, detail preservation, and overall visual quality.

\cref{fig:us_result} reports the per-method scores averaged across participants and the four style-count settings. {\MAST} achieves the highest overall mean score of $5.14/7$, ranking first on style faithfulness ($5.47$), structural consistency, detail preservation, and overall quality. Among the baselines, CoCoDiff~\cite{CoCoDiff} is the strongest overall ($4.80/7$), while StyleGallery~\cite{StyleGallery} attains the highest content-preservation score of $5.71$ yet the lowest style-faithfulness score of $3.06$, showing that content preservation alone does not imply successful style transfer.

\begin{figure}[t]
\centering
    \includegraphics[width=\linewidth]{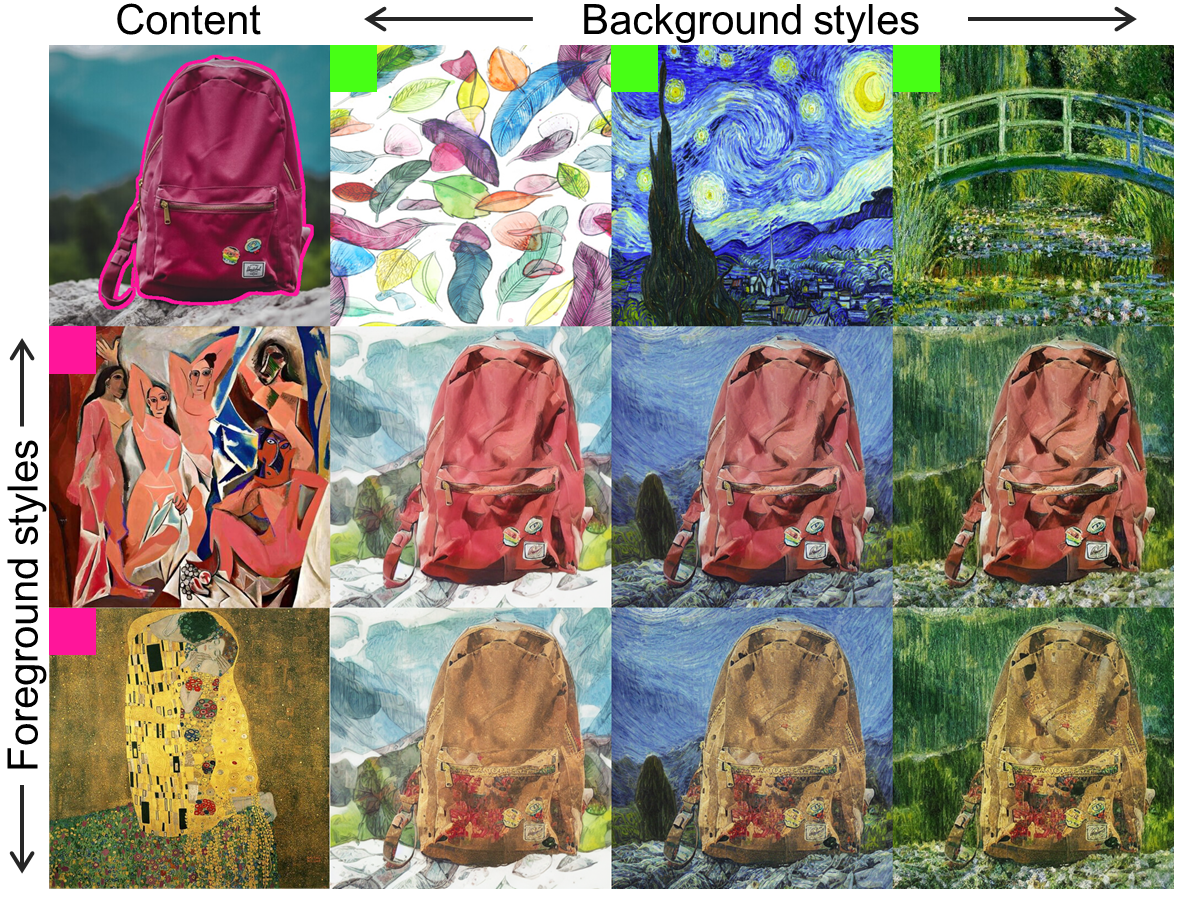}
    \caption{Compositional interaction in the two-style setting
    (\textcolor{green}{green}: background).}
    % Each row fixes the style
    % assigned to the foreground region while varying only the background
    % style. The assigned foreground style remains identifiable and
    % spatially localized, while its color and texture expression adapts to
    % the accompanying background. This demonstrates joint multi-style
    % integration rather than independent regional stylization.
\label{fig:inter}
\end{figure}

% 위에 틀 유지하면서, 아래 결과는 다르게 기술하기 
% competitive quality in the single-style setting, their scores tend to
% drop when extended to the multi-style setting. In contrast, our
% method performs even better in the multi-style setting, indicating
% that the proposed design effectively balances and harmonizes style
% and content, leading to more satisfactory results.

\noindent {\bf Robustness across style counts and spatial configurations.}
\cref{fig:app_ours_fig} qualitatively evaluates {\MAST} across one
to four style references, multiple image aspect ratios, and spatial
dimensions ranging from $512$ to $1024$ pixels, extending beyond the
fixed $512\times512$ setting used in the main experiments. Across these
configurations, the assigned styles remain localized to their target
regions, while the main content structures are retained and boundaries
between neighboring regions remain visually coherent.

Extending the progressive analysis up to four styles shown in
the main paper (\cref{fig:sequential_style}), \cref{fig:ab_5sty} introduces up
to five styles one at a time under a fixed spatial partition. Despite
the greater multi-style complexity and the presence of narrow mask
regions, regions stylized at earlier stages remain visually stable as
new references are introduced, while content structure and
foreground--background separation are maintained. In the dog and pizza
examples in the final two rows, the distinctive characteristics of the
references assigned to the orange and blue masks---including the
luminous appearance of the ice-castle reference and the vivid palette
of the cat reference---remain confined to their intended regions, with
little visible interference in neighboring areas. Together,
\cref{fig:app_ours_fig} and \cref{fig:ab_5sty} provide qualitative
evidence that {\MAST} remains effective across the tested reference
counts, aspect ratios, resolutions, and regional layouts.

\begin{figure*}[t]
\centering
    \includegraphics[width=\linewidth]{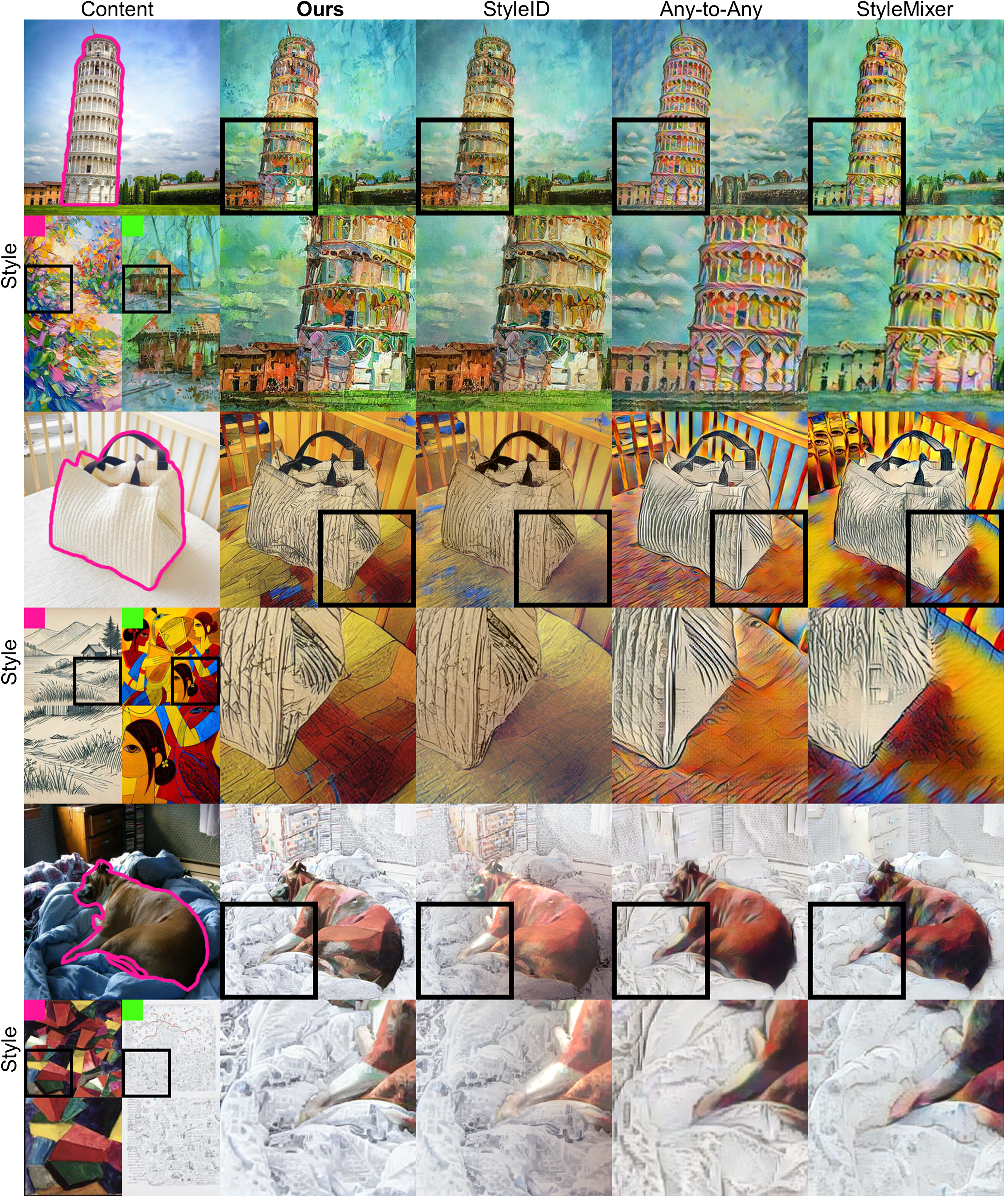}
    \caption{
    Regional style-fidelity comparison with the lowest-ArtFID \textbf{G-S}, \textbf{R-S},
    and \textbf{R-M} baselines in the two-style setting
    (\textcolor{green}{green}: background region). Enlarged views pair each
    style reference with its corresponding stylized region.
    }
    \label{fig:app_bo}
\end{figure*}

\noindent {\bf Compositional interaction in the two-style setting.}

The preceding results demonstrate the scalability of {\MAST}: adding
style references preserves the identity and spatial localization of
previously stylized regions. They do not, however, reveal whether these
regions are stylized independently or whether their local appearance
adapts to the surrounding style composition. We examine this
complementary question in the two-style setting.

\cref{fig:inter} isolates this interaction by fixing the foreground
style within each row while varying only the background reference.
Despite using the same foreground reference, the resulting foreground
changes in local color and texture across columns, while its characteristic
style remains recognizable and confined to the assigned region. Thus,
the variation reflects contextual adaptation to the accompanying style
rather than direct cross-style leakage.

This contextual adaptation arises naturally from the shared denoising
process. Although the foreground reference is fixed, changing the
background style alters the evolving stylization features and therefore
the content-anchored query used to retrieve foreground style features.
In addition, STS estimates a single sharpening factor for each attention
head from the query-averaged sharpness gap. A change in the background
attention distribution can thus modify the head-wise factor $\tau$,
which is subsequently applied to foreground queries as well. Consequently,
the same foreground reference can emphasize different subsets of its
style features depending on the surrounding style composition, while its
overall identity and spatial assignment remain preserved. This explains
the behavior in \cref{fig:inter} as contextual adaptation within a
joint multi-style generation process rather than direct cross-style
leakage.

\noindent {\bf Regional style fidelity in the two-style setting.}
Complementing the full-image comparison in the main paper (\cref{fig:exp_base}), \cref{fig:app_bo} provides enlarged
regional comparisons between {\MAST} and the best-performing baseline
from each of the three method categories in terms of ArtFID:
StyleID~\cite{StyleID} (\textbf{G-S}), Any-to-Any~\cite{Any-to-any} (\textbf{R-S}), and
StyleMixer~\cite{StyleMixer} (\textbf{R-M}). By pairing each style reference
with its corresponding stylized region, the figure enables direct
inspection of how well local texture patterns, color tones, and other
stylistic attributes are reproduced.

Across these examples, the baseline results show attenuated texture
patterns, deviations in color tone, or reduced stylistic distinctiveness
relative to their references. By contrast, {\MAST} more faithfully
reproduces the characteristic local attributes of both references while
maintaining visually coherent boundaries between adjacent regions.
This qualitative comparison indicates stronger style--region
correspondence in the two-style setting.

\end{document}